\documentclass[12pt, draftclsnofoot, onecolumn]{IEEEtran}
\usepackage{times}
\usepackage[final]{graphicx}
\usepackage{textcomp}
\usepackage[reqno]{amsmath}
\usepackage{amsfonts}
\usepackage{times,amsmath,epsfig}
\usepackage{latexsym,amssymb}
\usepackage{cite}
\usepackage{comment}
\usepackage{psfrag}
\usepackage{stfloats}
\usepackage{float}
\usepackage{amsmath}
\usepackage{amssymb}
\usepackage{array}
\usepackage[noend]{algorithmic}
\usepackage[ruled, linesnumbered]{algorithm2e}
\usepackage{bm}
\usepackage{color}
\usepackage{multicol}
\usepackage{multirow}
\usepackage{setspace}
\usepackage{tabularx}
\usepackage[T1]{fontenc}
\usepackage{subeqnarray}
\usepackage{cases}
\usepackage{makecell}

\newtheorem{lem}{\bf{Lemma}}
\newtheorem{prob}{\bf{Problem}}
\newtheorem{remark}{\bf{Remark}}
\usepackage{subfigure}
\usepackage{pdfpages}
\usepackage{graphicx}
\newlength{\figwidth}
\newcommand{\PreserveBackslash}[1]{\let \temp =\\#1 \let \\ = \temp}
\newcolumntype{C}[1]{>{\PreserveBackslash\centering}p{#1}}
\newcolumntype{R}[1]{>{\PreserveBackslash\raggedleft}p{#1}}
\newcolumntype{L}[1]{>{\PreserveBackslash\raggedright}p{#1}}

\newtheorem{assumption}{\bf{Assumption}}

\begin{document}

\title{Over-the-Air Computation Aided Federated Learning With the Aggregation of Normalized Gradient}

\author{
Rongfei Fan, Xuming An, Shiyuan Zuo, and Han Hu

\thanks
{
X. An, S. Zuo, and H. Hu are with the School of Information and Electronics, Beijing Institute of Technology, Beijing 100081, P. R. China. (\{3120195381,3120210836,hhu\}@bit.edu.cn).
}
\thanks{R.~Fan is with the School of Cyberspace Science and Technology, Beijing 100081, P. R. China. (fanrongfei@bit.edu.cn).}
}

\maketitle

\begin{abstract}
Over-the-air computation is a communication-efficient solution for federated learning (FL). In such a system, iterative procedure is performed: Local gradient of private loss function is updated, amplified and then transmitted by every mobile device; the server receives the aggregated gradient all-at-once, generates and then broadcasts updated model parameters to every mobile device.
In terms of amplification factor selection, most related works suppose the local gradient's maximal norm always happens although it actually fluctuates over iterations, which may degrade convergence performance. To circumvent this problem, we propose to turn local gradient to be normalized one before amplifying it. Under our proposed method,
when the loss function is smooth, we prove our proposed method can converge to stationary point at sub-linear rate.
In case of smooth and strongly convex loss function, we prove our proposed method can achieve minimal training loss at linear rate with any small positive tolerance. Moreover, a tradeoff between convergence rate and the tolerance is discovered.
To speedup convergence, problems optimizing system parameters are also formulated for above two cases.
Although being non-convex, optimal solution with polynomial complexity of the formulated problems are derived.
Experimental results show our proposed method can outperform benchmark methods on convergence performance.
\end{abstract}

\begin{IEEEkeywords}
Federated learning (FL), over-the-air computation, the aggregation of normalized gradient
\end{IEEEkeywords}

\section{Introduction}
In recent decade, intelligent applications are growing rapidly for mobile devices, which will impose a heavy traffic burden on network and lead to privacy leakage when performing model training \cite{FL1}. Federated learning is a new distributed machine learning framework that does not need to bring the raw data of any mobile device to a central server \cite{FL2}.
Specifically, in an FL system, all the involving mobile devices and the central server cooperatively interact over multiple rounds. In each round, every mobile device needs to generate the gradient of local loss function, which is abbreviated as {\em local gradient}, based on the received public model parameter in the last iteration, and then offloads its local gradient to the central server. Subsequently, the central server aggregates all the received local gradients to produce the public model parameter for the current iteration and then broadcast it to every mobile device \cite{FL3, FL4}.

Although the issue of offloading massive and private raw data has been avoided in an FL system, the aggregation of every mobile device's local gradients is still a heavy task as they have to be recovered by the central server one-by-one. To be more communication-efficient, the over-the-air computation technique can be adopted, which exploits the signal supposition property of a wireless multiple-access channel and allows simultaneous transmission of every mobile device's gradient embedding signal \cite{Air0}.
In this way, the central server can get the aggregated gradient all-at-once without decoding any specific mobile device's local gradient \cite{Air1}.

In an over-the-air computation-aided FL system, the local gradient to be transmitted at every mobile device and the received signal at the central server is usually amplified by an adjustable ratio to overcome the aggregation bias and enhance the training performance \cite{TWC_Poor, JSAC_Angela, Cui_JSAC, TCOM_Derrick, JSAC_Niu}.
However, due to the fluctuation of the local gradient over iterations (as verified in our experiment results) and the limited transmit power of every mobile device, many existing works take the most conservative assumption on the norm of the local gradient by adopting its maximal value \cite{TWC_Poor,JSAC_Angela, Cui_JSAC,TCOM_Derrick,JSAC_Niu}.
This will lead to the shrinkage of the signal amplification factor's feasible region, especially when the local gradient trends to be zero, which would be deleterious for suppressing the training loss.
Some literatures have seen the above limitation and try to flatten local gradient's norm over iterations \cite{TWC_Huang,Angle_RIS_TWC}.
Specifically, only the sign (+1 or -1) of every element of local gradient is transmitted in \cite{TWC_Huang}, which promises the transmitted signal before amplification to be with constant amplitude; the local gradient vector is firstly deducted by the vector's mean and then divided by the vector's standard deviation before amplification in \cite{Angle_RIS_TWC}, which makes the transmitted signal before amplification to be with zero mean and unit variance.

With regard to the above two efforts, although the norm of local gradient over iterations has been flattened, the convergence performance of the method in \cite{TWC_Huang} may be degraded due to the great loss of gradient information (which has been verified in experiments of this paper), and the operation of \cite{Angle_RIS_TWC} cannot guarantee the transmitted signal before amplification to be bounded by some certain value, which still poses uncertainty to the transmitter.
In contrast, we can use another norm flattening method: Normalize the local gradient before amplification, i.e., divide every element of the local gradient vector by its norm. Through this operation, not only the norm of local gradient is fixed at one, but also the absolute value of every element of local gradient is always bounded by one.

\subsection{Related Works} \label{e:related_works}

In related literature on over-the-air aided FL system, the efforts in early stage focus on minimizing the MSE for gradient aggregation only through the optimization of signal amplification factors at every mobile device and the ES \cite{Air2,Air3,MSE3,Air4}, which is thought be highly related to training loss.
In \cite{TSP_Cohen}, rather than minimizing the MSE of gradient aggregation, the gap to minimal training loss is characterized for the first time when the loss function is smooth only or with strong convexity as well, without adjusting any signal amplification factor.

Hereafter, the issue of adjusting signal amplification factor is taken into account in the literature characterizing the gap to minimal training loss, due to the restriction of maximal allowable transmit power of every mobile device and for speeding up convergence, under various scenarios.
In \cite{TWC_Poor} and  \cite{JSAC_Angela}, signal amplification factor is carefully selected to obey the limit of maximal transmit power for associated mobile device, with a consideration of blind channel station information and sparse local gradient, respectively.
Some other works optimize, rather than just select, signal amplification factor so as to achieve better training performance \cite{Cui_JSAC,TCOM_Derrick,JSAC_Niu}.
\cite{Cui_JSAC} considers a traditional over-the-air aided FL system;
\cite{TCOM_Derrick} investigates a multiple-input multiple-output (MIMO) system with analog beamforming at every mobile device;
\cite{JSAC_Niu} imposes energy consumption budget for every mobile device in each iteration.

On the other hand, when it comes to the setting of signal amplification factor for mobile device, due to the uncertainty of local gradient over iterations, most conservative assumption is made on local gradient's norm by taking its maximal value in aforementioned works, which restricts the feasible region of signal amplification factor and affects convergence performance. To overcome this problem,
\cite{TWC_Huang} proposes to transmit one-bit information of local gradient, and \cite{Angle_RIS_TWC} uniforms the local gradient vector with its every element subtracted with the vector's mean and then divided with the its standard deviation.

It should be highlighted that the gap to minimal training loss (or stationary point in case of non-convex loss function) in aforementioned works can seldom reach to zero as the number of iteration round grows.
Especially, when the loss function is smooth only, no existing work has shown convergence result.
When the loss function is not only smooth but also strongly convex (or satisfies Polyak-{\L}ojasiewicz inequality, which will hold when the loss function is strongly convex), both \cite{Cui_JSAC} and \cite{Angle_RIS_TWC}  have claimed the convergence to an $\varepsilon$-gap to minimal training loss at a linear rate.

\subsection{Contributions and Outline}
Motivated by the above, this work proposes a new gradient aggregation method, which normalizes local gradients before amplification.
Under such a newly proposed aggregation method, convergence performance is investigated for various types of loss functions by characterizing the associated training loss. System parameters are subsequently optimized so as to minimize the derived training losses.
The main contributions of this paper are summarized as follows:

\begin{itemize}
\item {\textbf{Proposing to aggregate normalized gradient:} We propose a method that normalizes the local gradient at every mobile device before amplifying
it. Compared with aggregating the local gradients directly, the proposed method can mitigate the negative impact of the conservative assumption on local gradient's norm, which restricts the feasible region of signal amplification factor and affects the convergence performance.}
\item {\textbf{Achieving stationary point for smooth only loss function at sub-linear rate:}} When the loss function is smooth only, we prove that our proposed method can achieve a stationary point of loss function as the number of iteration goes to infinity at a sub-linear rate, which has never been claimed in the literature on over-the-air computation aided FL to the best our knowledge.
\item \textbf{Achieving $\varepsilon$-small optimality gap for both smooth and strongly convex loss function at linear rate:} When the loss function is not only smooth but also strongly convex, with the support of system parameter optimization, we prove our proposed method can converge to any small $\varepsilon$-gap to the minimal training loss as the number of iteration goes to infinity at a linear rate. Furthermore, a tradeoff between the convergence rate and $\varepsilon$-gap is discovered.
\item \textbf{Optimally solving non-convex parameter optimization problems:} In terms of system parameter optimization for the above two cases, non-convex optimization problems are involved, whose optimal solutions are hard to find. With mathematical transformation and analysis, we transform them to be a combination of bisection search and convex problem equivalently, whose optimal solution is achievable in polynomial time.
\item \textbf{Experiment results: } The numerical reuslts validate the performance of proposed method. Under the same settting, the proposed method can achieve higher test accuracy and faster convergence speed than the two benchmarks in cases that the loss fucntion is convex or non-convex. 
\end{itemize}

The rest of this paper is organized as follows. The system model is given in Section \ref{s:model}. Convergence analysis is shown in Section \ref{s:convergence}. Minimization of FL training loss is performed in Section \ref{s:optimization}. Experimental results are presented in Section \ref{s:exp}, followed by concluding remarks in Section \ref{s:conclu}.

\section{System Model}  \label{s:model}
Consider a FL system with $K$ mobile devices and one edge server (ES), which has a wireless link with every mobile device.
These $K$ mobile devices compose the set of $\mathcal{K} \triangleq \{1, 2, ..., K\}$.
For any mobile device, say $k$th mobile device, $k \in \mathcal{K}$, it has a local data set $\mathcal{D}_k$ with $D_k$ elements.  The $i$th element of $\mathcal{D}_k$ is a ground-true label $\{\bm{x}_{k,i}, \bm{y}_{k,i}\}$, for any $i \in \mathcal{D}_k$.
The $\bm{x}_{k,i} \in \mathcal{R}^{\text{in}}$ is the input vector and $\bm{y}_{k,i} \in \mathcal{R}^{\text{out}}$ is the output vector.
By utilizing the data set  $\mathcal{D}_k$ for $k\in \mathcal{K}$ and with the coordination of these $K$ mobile devices and the ES, and define $D_A = \sum_{k\in \mathcal{K}} D_k$,
a machine learning task is aimed to be completed to train a $N$ dimension vector $\bm{w}$ by minimizing the following loss function, which is also called as training loss in the sequel,
\begin{equation}
F(\bm{w}) \triangleq  \frac{1}{D_A} \sum_{k\in \mathcal{K}} \sum_{i\in \mathcal{D}_k} f(\bm{w}, \bm{x}_{k,i}, \bm{y}_{k,i})
=  \sum_{k\in \mathcal{K}} \frac{D_k}{D_A} F_k(\bm{w})
\end{equation}
where $f(\bm{w}, \bm{x}_{k,i}, \bm{y}_{k,i})$ is the loss function to evaluate the error for approximating the $\bm{y}_{k,i}$ with an input of $\bm{x}_{k,i}$ and a selection of $\bm{w}$, and $F_k(\bm{w})$ represents the local loss function of $k$th mobile device and can be defined as
\begin{equation}
F_k(\bm{w}) \triangleq \frac{1}{D_k} \sum_{i\in \mathcal{D}_k} f(\bm{w}, \bm{x}_{k,i}, \bm{y}_{k,i}), \forall k \in \mathcal{K}.
\end{equation}

For the defined loss function $F(\bm{w})$ and local loss function $F_k(\bm{w})$, as shown not only in literature on over-the-air computation aided FL \cite{TWC_Poor,JSAC_Angela, Cui_JSAC,TCOM_Derrick,JSAC_Niu,TWC_Huang, Angle_RIS_TWC} but also in literature purely on FL \cite{Pure_FL_1,Pure_FL_2,Pure_FL_3,Pure_FL_4,Pure_FL_5}, one or more assumptions can be imposed on them.
\begin{assumption}[Smoothness]\label{a:smooth}
Let $\nabla F(\bm{w})$ denote the gradient vector of the loss function $F(\bm{w})$ at the point of $\bm{w}$, then there is non-negative constant $L$ such that
\begin{equation}\label{g-smooth}
    \lVert \nabla F(\bm{w}_1) - \nabla F(\bm{w}_2) \rVert \le L \lVert \bm{w}_1-\bm{w}_2 \rVert , \forall \bm{w}_1, \bm{w}_2 \in \mathbb{R}^N.
\end{equation}
\end{assumption}

\begin{assumption}[Strong Convexity] \label{a:convexity}
The loss function $F(\bm{w})$ is $M$-strongly convex $(M>0)$, i.e., the following inequality
\begin{equation}
\langle \nabla F(\bm{w}_1) - \nabla F(\bm{w}_2), \bm{w}_1-\bm{w}_2 \rangle \geq M \lVert \bm{w}_1 - \bm{w}_2 \rVert ^2
\end{equation}
holds for any $\bm{w}_1, \bm{w}_2 \in \mathcal{R}^{N}$.
\end{assumption}

\begin{assumption}[Bounded Gradient Norm] \label{a:gradient_bound}
The norm of $F_k(\bm{w})$'s gradient is upper-bounded by $G$ for $k\in \mathcal{K}$, i.e.,
\begin{equation}
\lVert \nabla F_k(\bm{w}) \rVert \leq G, \forall k\in \mathcal{K}, \bm{w} \in \mathcal{R}^{N},
\end{equation}
which implies that
\begin{equation}
\lVert \nabla F(\bm{w}) \rVert \leq \sum_{k\in \mathcal{K}} \frac{D_k}{D_A} \lVert \nabla F_k(\bm{w}) \rVert \leq G,
\end{equation}
since the norm function $\lVert \cdot \rVert$ is convex.
\end{assumption}

Due to statistical heterogeneity of ground-true labels at multiple mobile devices, the evaluated local loss function $F_k(\bm{w})$ over $k\in \mathcal{K}$ will be different.
Compared with $F(\bm{w}) = \sum_{k\in \mathcal{K}} \left(\frac{D_k}{D_A}\right) F_k(\bm{w})$, which represents the mean of all the local loss functions over the global ground-true labels, each specific $F_k(\bm{w})$ for $k\in \mathcal{K}$ can be taken as a random variable. Two mild assumptions on the statistical characters of $F_k(\bm{w})$ for $k\in \mathcal{K}$ are given as follows
:
\begin{assumption}[Independent Distribution] \label{a:independent}
For any pair of $k, k' \in \mathcal{K}$ and $k\neq k'$, $F_k(\bm{w})$ and $F_{k'}(\bm{w})$ (and thus $\nabla F_k(\bm{w})$ and $\nabla F_{k'}(\bm{w})$, or $\frac{\nabla F_k(\bm{w})}{\lVert \nabla F_{k}(\bm{w}) \rVert}$ and $\frac{\nabla F_{k'}(\bm{w})}{\lVert \nabla F_{k'}(\bm{w}) \rVert}$) are independently distributed from each other, $\forall \bm{w} \in \mathcal{R}^N$.
\end{assumption}

\begin{assumption}[Limited Bias] \label{a:bias}
For any $k\in \mathcal{K}$, the bias between $\nabla F_k(\bm{w})$ and $\nabla F(\bm{w})$, which is measured by the angle between the two vectors $\nabla F_k(\bm{w})$ and $\nabla F(\bm{w})$, is limited, i.e., the
\begin{equation}
\theta_k(\bm{w}) \triangleq \cos^{-1} \left(\frac{\nabla F(\bm{w})^T \nabla F_k(\bm{w})}{\lVert F(\bm{w}) \rVert \lVert \nabla F_k(\bm{w}) \rVert} \right), \forall k \in \mathcal{K},
\end{equation}
satisfies such an inequality
\begin{equation}
|\theta_k(\bm{w})| \leq \theta_{\text{th}}, \forall k \in \mathcal{K}, \forall \bm{w} \in \mathcal{R}^N,
\end{equation}
where $0< \theta_{\text{th}} < \pi/2$
.
\end{assumption}


Traditionally in a FL system, three steps of operations are performed iteratively until convergence.
In $t$th iteration, these three steps are as follows:
\begin{itemize}
\item Step 1 (Local Update): Each mobile device, say $k$th mobile device, calculates the gradient of its local loss function $F_k(\bm{w})$ at the point $\bm{w}=\bm{w}^{(t)}$, which can be also written as $\bm{g}_{k}^{(t)} = \nabla F_k(\bm{w}^{(t)})$. The $\bm{w}^{(t)}$ represents the commonly shared $\bm{w}$ at the end of $(t-1)$th iteration.
\item Step 2 (Aggregation): Each mobile device uploads the associated $\bm{g}_{k}^{(t)}$ to the ES. Then the ES aggregates the $\{\bm{g}_{k}^{(t)}| k \in \mathcal{K}\}$ to generate $\bm{w}^{(t+1)}$. One broadly used aggregation method is given as
\begin{equation}
\bm{w}^{(t+1)} = \bm{w}^{(t)} - \eta^{(t)} \sum_{k\in \mathcal{K}} \frac{D_k}{D_A} \bm{g}_k^{(t)}, \forall t = 1, 2, ...
\end{equation}
where $\eta^{(t)}$ is the learning rate in $t$th iteration.
\item Step 3 (Broadcast): The ES broadcasts $\bm{w}^{(t+1)}$ to every mobile device.
\end{itemize}


In Step 2 (Aggregation) of every iteration, there are multiple mobile devices trying to upload their local gradients $\bm{g}_k^{(t)}$ to the ES through a multiple access control (MAC) channel. To be computation-efficient, over-the-air computation technique is adopted.
Suppose the signal to be amplified and then transmitted by the $k$th mobile device in $t$th iteration is $\bm{x}_k^{(t)}$, denote the signal amplification factor at the side of mobile device as $b_k$, which is no larger than $b_k^{\max}$ due to the limit of maximal transmit power, and the signal amplification factor at the side of ES as $a$ \footnote{There is no upper bound imposed on $a$ because the received signal at the side of ES can be firstly quantized to be a digital signal and then scaled up by any ratio.}, assume the channel coefficient between the $k$th mobile device to the ES in $t$th iteration is $h_k$, then the received signal in the $t$th iteration at the ES, denoted as $\bm{y}^{(t)}$, can be written as
\begin{equation}
\bm{y}^{(t)} = a \left(\sum_{k\in \mathcal{K}} \bm{x}_k^{(t)} {b_k} h_k + \bm{z}^{(t)}\right), \forall t = 1, 2, ...
\end{equation}
where $\bm{z}^{(t)}$ is additive Gaussian noise vector with mean being $\bm{0}$ and variance being $\sigma^2 \bm{I}$.
Then the aggregation method in Step 2 (Aggregation) can be updated as
\begin{equation}
\bm{w}^{(t+1)} = \bm{w}^{(t)} - \eta^{(t)} a \left( \sum_{k\in \mathcal{K}} \bm{x}_k^{(t)} {b_k} h_k + \bm{z}^{(t)}\right), \forall t = 1, 2, ...
\end{equation}
{In existing literatures on over-the-air computation aided FL system, $\bm{x}_k^{(t)}$ is usually selected to be $\bm{g}_k^{(t)}$ for $k\in \mathcal{K}$\cite{Cui_JSAC,JSAC_Niu,JSAC_Angela}}.
In this paper,  differently, we set $\bm{x}_k^{(t)}$ to be the normalized gradient of $\bm{g}_k^{(t)}$, i.e.,
\begin{equation}
\bm{x}_k^{(t)} \triangleq {\bm{g}_k^{(t)}}/{\lVert \bm{g}_k^{(t)} \rVert}, \forall k \in \mathcal{K}, \forall t = 1, 2, ...
\end{equation}

Under the above formulated system model, convergence performance under our proposed aggregation method will be analyzed in Section \ref{s:convergence} and improved by optimizing the system parameters in Section \ref{s:optimization}, respectively.

\section{Convergence Analysis} \label{s:convergence}

\subsection{Case I: With Smoothness Only for Loss Function}
In this part, we merely impose Assumption \ref{a:smooth} (Smoothness) on loss function $F(\bm{w})$. With smoothness only, $F(\bm{w})$ may be non-convex and the point of $\bm{w}$ such that $\nabla F(\bm{w}) = 0$ can be taken as the stationary point of $F(\bm{w})$ (rather than the global optimal solution for minimizing $F(\bm{w})$) and is defined as $\bm{w}^*$ in subsequent discussion.
\begin{lem} \label{lem:CaseI}
By setting learning rate $\eta^{(t)}$ as $\eta^{(t)} = {1}/{t^p}$ with ${1}/{2} < p < 1$ and selecting $a>0$ and $\{b_k| k\in \mathcal{K}\}$ such that $\sum_{k\in \mathcal{K}} h_k b_k >0$, the term $\mathop{\min} \limits_{t \in [0,T]}  \lVert \nabla F(\bm{w}^t) \rVert $'s upper bound is given in (\ref{e:L_smooth_lemma}) and will converge to zero at a sub-linear rate as $T$ grows to infinity.
\begin{align} \label{e:L_smooth_lemma}
& \mathop{\min} \limits_{t \in [0,T]}  \lVert \nabla F(\bm{w}^{(t)}) \rVert  \notag \\
 \leq &
 \frac{\mathbb{E} \left\{F(\bm{w}^{(1)}) - F(\bm{w}^{(T+1)})\right\}}{  T^{1- p} \cdot \cos (\theta_{\text{th}})  a \sum_{k\in \mathcal{K}} h_k b_k } \notag \\
 & + \frac{2p}{T^{1-p} \left(2p-1\right)} \times
 \left(\frac{a L }{2 \cos (\theta_{\text{th}})  \sum_{k\in \mathcal{K}} h_k b_k} \right) \times
 \left( \sum_{k\in \mathcal{K}} 4 h_k^2 b_k^2 + \left(\sum_{k\in \mathcal{K}} h_k b_k \right)^2 + n \sigma^2 \right)
\end{align}
\end{lem}
\begin{IEEEproof}
Please refer to Appendix \ref{app:lemma1}.
\end{IEEEproof}
\begin{remark} \label{r:CaseI}
With Lemma \ref{lem:CaseI},
the stationary point of $F(\bm{w})$ can be achieved at a sub-linear rate as $T$ grows to infinity.
\end{remark}
\subsection{Case II: With Smoothness and Strong Convexity for Loss Function}
In this part, we impose not only Assumption \ref{a:smooth} (Smoothness) but also Assumption \ref{a:convexity} (Strong Convexity) on loss function $F(\bm{w})$. In such a case,
the following lemma can be anticipated to characterize the associated optimality gap.

\begin{lem} \label{lem:strong_convex}
By setting learning rate $\eta^{(t)}$ as $\eta>0$, selecting $a>0$ and $\{b_k| k\in \mathcal{K}\}$ such that $\sum_{k\in \mathcal{K}} h_k b_k >0$, and defining
\begin{equation}
q^{\max} \triangleq  \max \left(\left(1 - \frac{2M \cos (\theta_{\text{th}})\eta a \sum_{k\in \mathcal{K}} h_k b_k}{G}\right), 0\right),
\end{equation}
the optimality gap $\left(F(\bm{w}^{(t+1)}) - F(\bm{w}^*)\right)$ is bounded in (\ref{e:Lemma2_ineq}) as follows
\begin{align} \label{e:Lemma2_ineq}
F(\bm{w}^{(T)}) - F(\bm{w}^*) \leq & \frac{L}{2} \left(q^{\max}\right)^{T-1}  \lVert \bm{w}^1 - \bm{w}^*\rVert^2 \notag \\
& + \frac{L}{2}  \max\left(\frac{a \eta G}{2M \cos (\theta_{\text{th}}) \sum_{k\in \mathcal{K}} h_k b_k} ,a^2 \eta^2 \right)
\times \left(\sum_{k\in \mathcal{K}} 4 h_k^2 b_k^2 + \left(\sum_{k\in \mathcal{K}} h_k b_k \right)^2 + n \sigma^2 \right).
\end{align}
\end{lem}
\begin{IEEEproof}
Please refer to Appendix \ref{app:lemma2}.
\end{IEEEproof}

In terms of the convergence performance for Case II, including whether the optimality gap can converge to zero and the associated convergence speed, the answer will be only clear after optimizing the adjustable parameters, including $a$, $\eta$, and $\{b_k | k\in \mathcal{K}\}$, to minimize the derived bound in (\ref{e:Lemma2_ineq}), which will be shown in Section \ref{s:optimization_CaseII}.
Hence concluding remarks on convergence performance for Case II will be disclosed at the end of Section \ref{s:optimization}, as given in Remark \ref{r:CaseII}.

\section{Minimization of FL Training Loss} \label{s:optimization}

In this section, in order to speedup convergence, the training loss is minimized through optimizing the adjustable system parameters, including $a$, $\{b_k| k\in \mathcal{K}\}$ and $\eta$, for Case I and Case II, respectively.

\subsection{Optimization for Case I} \label{s:optimization_CaseI}
In this case, as shown in (\ref{e:L_smooth_lemma}), the characterized upper bound has two parts related to $a$ and $\{b_k | k \in \mathcal{K}\}$, which are given as
\begin{equation}
C_{1}^{\text{I}} \left(a, \{b_k|k\in \mathcal{K}\} \right) \triangleq \frac{1}{a \sum_{k\in \mathcal{K}} h_k b_k },
\end{equation}
and
\begin{equation}
\begin{array}{ll}
 C_{2}^{\text{I}} \left(a, \{b_k|k\in \mathcal{K}\} \right)
 \triangleq \frac{a}{\sum_{k\in \mathcal{K}} h_k b_k } \bigg(\sum_{k\in \mathcal{K}} 4 h_k^2 b_k^2
 + \left(\sum_{k\in \mathcal{K}} h_k b_k \right)^2 + n \sigma^2 \bigg).
\end{array}
\end{equation}
To suppress the training loss, both $C_{1}^{\text{I}} \left(a, \{b_k|k\in \mathcal{K}\} \right)$ and $C_{2}^{\text{I}} \left(a, \{b_k|k\in \mathcal{K}\} \right)$ are required to be compressed, which is a conflicting task since $C_{1}^{\text{I}} \left(a, \{b_k|k\in \mathcal{K}\} \right)$ is monotonically decreasing with $a$ and $\{b_k| k\in \mathcal{K}\}$, while $C_{2}^{\text{I}} \left(a, \{b_k|k\in \mathcal{K}\} \right)$ is monotonically increasing with $a$ and has uncertain monotonicity with $\{b_k | k\in \mathcal{K}\}$.

Hence both $a$ and $\{b_k | k\in \mathcal{K} \}$ should be properly set so as to achieve the best tradeoff between
$C_{1}^{\text{I}} \left(a, \{b_k|k\in \mathcal{K}\} \right)$ and $C_{2}^{\text{I}} \left(a, \{b_k|k\in \mathcal{K}\} \right)$. Correspondingly, by setting $C_{1}^{\text{I}} \left(a, \{b_k|k\in \mathcal{K}\} \right)$ as $S$, the following optimization problem needs to be solved
\begin{prob} \label{p:CaseI_lower}
\begin{subequations}
\begin{align}
 C_2^{\text{I}}(S) \triangleq \mathop{\min} \limits_{a,  \{b_k | k\in \mathcal{K}\}} \quad & C_{2}^{\text{I}} \left(a, \{b_k|k\in \mathcal{K}\} \right)   \nonumber\\
  \text{s.t.}  \quad & C_{1}^{\text{I}} \left(a, \{b_k|k\in \mathcal{K}\} \right) = S, \label{e:CaseI_lower_a} \\
  			   \quad & a > 0, \label{e:p1stlk}\\
  			   \quad & 0 \leq b_k \leq b_k^{\max}, \forall k \in \mathcal{K}, \\
			   \quad & \sum_{k\in \mathcal{K}} h_k b_k >0.	\label{e:CaseI_lower_d}	
  			   \end{align}
\end{subequations}
\end{prob}

With such a definition, the training loss minimization problem can be solved by selecting a proper value of $S>0$ so as to minimize the following cost function,
\begin{equation}
\frac{\left( {S \mathbb{E} \left \{F(\bm{w}^{(1)}) - F(\bm{w}^{(T+1)})\right \}}
+  \frac{L p}{ (2p-1)} C_2^{\text{I}}(S) \right)}{T^{1-p} \cos(\theta_{\text{th}})},
\end{equation}
which is equivalent with the following optimization problem
\begin{prob} \label{p:CaseI_upper}
\begin{subequations}
\begin{align}
\mathop{\min} \limits_{S}  \quad &  {S \mathbb{E} \left \{F(\bm{w}^{(1)}) - F(\bm{w}^{(T+1)})\right \}} + \frac{L p}{\left(2p-1\right)}  C_2^{\text{I}}(S) \notag \\
 \text{s.t.}  \quad & S > 0.
\end{align}
\end{subequations}
\end{prob}

For Problem \ref{p:CaseI_lower}, by replacing $a= \frac{1}{S \sum_{k\in \mathcal{K}} h_k b_k}$ according to (\ref{e:CaseI_lower_a}), the objective function of Problem \ref{p:CaseI_lower} turns to be
\begin{equation} \label{e:C_2_I_full}
\frac{1}{S \left(\sum_{k\in \mathcal{K}} h_k b_k \right)^2 } \left(\sum_{k\in \mathcal{K}} 4 h_k^2 b_k^2  + \left(\sum_{k\in \mathcal{K}} h_k b_k \right)^2 +  n \sigma^2 \right).
\end{equation}
By omitting the constant value item ${\left(\sum_{k\in \mathcal{K}} h_k b_k \right)^2}/{\left(\sum_{k\in \mathcal{K}} h_k b_k \right)^2}$ and coefficient ${1}/{S}$, Problem \ref{p:CaseI_lower} can be equivalently reformulated as the following one \footnote{Note that in Problem \ref{p:CaseI_lower_recast}, the constraint (\ref{e:CaseI_lower_d}) in Problem \ref{p:CaseI_lower} is dropped because the case that $\sum_{k\in \mathcal{K}} h_k b_k =0 $ can lead to infinity of the cost function of Problem \ref{p:CaseI_lower_recast}, which will never happen at its optimal solution. In other words, minimizing the cost function of Problem \ref{p:CaseI_lower_recast} can naturally preclude the happening of $\sum_{k\in \mathcal{K}} h_k b_k =0 $.}
\begin{prob} \label{p:CaseI_lower_recast}
\begin{subequations}
\begin{align}
Z \triangleq \mathop{\min} \limits_{\{b_k | k\in \mathcal{K}\}} \quad & \frac{1}{\left(\sum_{k\in \mathcal{K}} h_k b_k \right)^2} \left(\sum_{k\in \mathcal{K}} 4 h_k^2 b_k^2  + n \sigma^2 \right)   \nonumber\\
  \text{s.t.}  \quad & 0 \leq b_k \leq b_k^{\max}, \forall k \in \mathcal{K}. \label{e:p1strk}		
\end{align}
\end{subequations}
\end{prob}

For Problem \ref{p:CaseI_lower_recast}, it can be checked to be a non-convex optimization problem since its objective function is non-convex with the vector $\{b_k | k\in \mathcal{K}\}$. Hence the optimal solution of Problem \ref{p:CaseI_lower_recast} can be hardly achieved in general.
To overcome this challenge, we transform Problem \ref{p:CaseI_lower_recast} into the following equivalent form
\begin{prob} \label{p:CaseI_lower_convex}
\begin{subequations}
\begin{align}
\mathop{\min} \limits_{r, \{b_k | k\in \mathcal{K}\}} \quad & r^2   \nonumber\\
  \text{s.t.}  \quad & \frac{1}{\left(\sum_{k\in \mathcal{K}} h_k b_k \right)^2} \left(\sum_{k\in \mathcal{K}} 4 h_k^2 b_k^2  +  n \sigma^2 \right) \leq r^2, \label{e:CaseI_lower_convex_a} \\
  \quad & 0 \leq b_k \leq b_k^{\max}, \forall k \in \mathcal{K},
\end{align}
\end{subequations}
\end{prob}
which is further equivalent with
\begin{prob} \label{p:CaseI_lower_convex_simple}
\begin{subequations}
\begin{align}
\mathop{\min} \limits_{{r}, \{b_k | k\in \mathcal{K}\}} \quad & {r}   \nonumber\\
 \text{s.t.}  \quad & \sqrt{ \left(\sum_{k\in \mathcal{K}} 4 h_k^2 b_k^2  + n \sigma^2 \right)} \leq {r} {\left(\sum_{k\in \mathcal{K}} h_k b_k \right)}, \label{e:CaseI_lower_convex_simple_a}\\
  \quad & 0 \leq b_k \leq b_k^{\max}, \forall k \in \mathcal{K}, \label{e:CaseI_lower_convex_simple_b} \\
  \quad & r >0.
\end{align}
\end{subequations}
\end{prob}

For Problem \ref{p:CaseI_lower_convex}, the following lemma can be expected.
\begin{lem} \label{lem:CaseI_convex_set}
With the $r$ given, the constraint in (\ref{e:CaseI_lower_convex_simple_a}) defines a convex set for the vector of $\{b_k|k\in \mathcal{K}\}$.
\end{lem}
\begin{IEEEproof}
Please refer to Appendix \ref{app:lemma3}.
\end{IEEEproof}

With the aid of Lemma \ref{lem:CaseI_convex_set}, by fixing $r$, Problem \ref{p:CaseI_lower_convex_simple} turns to be a convex optimization problem with the vector of $\{b_k | k\in \mathcal{K}\}$, which would be feasible if ${ \sqrt{ \left(\sum_{k\in \mathcal{K}} 4 h_k^2 b_k^2  + n \sigma^2 \right)}}/ {{\left(\sum_{k\in \mathcal{K}} h_k b_k \right)}}$ under the constraint  (\ref{e:CaseI_lower_convex_simple_b}) can achieve the current $r$
and would not be feasible otherwise.
Since Problem \ref{p:CaseI_lower_convex_simple}  in this case (with $r$ fixed) has no objective function, to be more operative, the feasibility check problem for Problem \ref{p:CaseI_lower_convex_simple} with a given $r$ can be reformulated as the following optimization problem
\begin{prob} \label{p:CaseI_lower_convex_feasibility}
\begin{subequations}
\begin{align}
V(r) \triangleq \mathop{\min} \limits_{v, \{b_k | k\in \mathcal{K}\}} \quad & {v}   \nonumber\\
 \text{s.t.}  \quad & \sqrt{ \left(\sum_{k\in \mathcal{K}} 4 h_k^2 b_k^2  + n \sigma^2 \right)} \leq {r} {\left(\sum_{k\in \mathcal{K}} h_k b_k \right)}, \label{e:CaseI_lower_convex_feasibility_a}\\
  \quad & 0 \leq b_k \leq b_k^{\max} + v, \forall k \in \mathcal{K} \label{e:CaseI_lower_convex_feasibility_b}.
\end{align}
\end{subequations}
\end{prob}
With the aid of Lemma \ref{lem:CaseI_convex_set}, it can be checked that Problem \ref{p:CaseI_lower_convex_feasibility} is also a convex optimization problem, which can be solved optimally by existing numerical methods in polynomial time at the order of $O((K+1)^3)$ \cite{Boyd}.
Moreover, according to the definition of Problem \ref{p:CaseI_lower_convex_feasibility}, when $V(r) \leq 0$, Problem \ref{p:CaseI_lower_convex_simple} with current input of $r$ is feasible. Otherwise, Problem \ref{p:CaseI_lower_convex_simple} with current input of $r$ is not feasible.

With the above operation, for a given $r$, we can check whether the term
${ \sqrt{ \left(\sum_{k\in \mathcal{K}} 4 h_k^2 b_k^2  + n \sigma^2 \right)}} / {{\left(\sum_{k\in \mathcal{K}} h_k b_k \right)}}$ under the constraint  (\ref{e:CaseI_lower_convex_simple_b}) can achieve the $r$, which also means the term  ${\left(\sum_{k\in \mathcal{K}} 4 h_k^2 b_k^2  + n \sigma^2 \right)}/{\left(\sum_{k\in \mathcal{K}} h_k b_k \right)^2}$ can achieve $r^2$.
To find the maximal ${\left(\sum_{k\in \mathcal{K}} 4 h_k^2 b_k^2  + n \sigma^2 \right)}/{\left(\sum_{k\in \mathcal{K}} h_k b_k \right)^2}$ under the constraint (\ref{e:CaseI_lower_convex_simple_b}), we only need to perform a bi-section search of $r$ to find the minimal $r$ such that $V(r)\leq 0$. To this end, Problem \ref{p:CaseI_lower_recast} has been solved optimally.

In the ending part of this subsection, we come to solve Problem \ref{p:CaseI_upper} by selecting a proper $S$.
Since the minimal cost function of Problem \ref{p:CaseI_lower_recast} is defined as $Z$, which is independent of $S$, then it can be found that $Z>0$ and $C_2^{\text{I}}(S) = {(Z+1)}/{S}$ according to the expression in (\ref{e:C_2_I_full}). Hence the optimal solution of $S$ for solving Problem \ref{p:CaseI_upper} can be obtained by setting the derivative of Problem \ref{p:CaseI_upper}'s cost function to be zero, which produces
\begin{equation} \label{e:case_I_S}
S = \sqrt{\frac{L (Z+1) p }{ (2p-1) \mathbb{E} \left \{F(\bm{w}^{(1)}) - F(\bm{w}^{(T+1)})\right \}}}.
\end{equation}
In real application, the value of $\mathbb{E}\left \{F(\bm{w}^{(1)}) - F(\bm{w}^{(T+1)})\right\}$ can be estimated from historical training results. Even if the information about $\mathbb{E}\left \{F(\bm{w}^{(1)}) - F(\bm{w}^{(T+1)})\right\}$ is not available, it is still meaningful to solve Problem \ref{p:CaseI_lower}, or equivalently Problem \ref{p:CaseI_lower_recast}, so as to suppress the training loss for a selected input of $S$.

At last, the algorithm to solve Problem \ref{p:CaseI_upper} for Case I is summarized in Algorithm \ref{a:case_I} as follows
\begin{algorithm}
\caption{The Procedure for Optimally Solving Problem \ref{p:CaseI_upper}}
\label{a:case_I}
\begin{algorithmic}[1]
\STATE \%\% Part I: Evaluate $Z$ for solving Problem \ref{p:CaseI_lower_recast}
\STATE Perform bi-section search of $r$ to find the minimal $r$, denoted as $r^*$, such that $V(r) \leq 0$. The evaluation of $V(r)$ is accomplished by solving Problem \ref{p:CaseI_lower_convex_feasibility}, which is a convex optimization problem.
\STATE Record the optimal $\{b_k\}$, denoted as $\{b_k^*\}$, when solving Problem \ref{p:CaseI_lower_convex_feasibility} for evaluating $V(r^*)$.
\STATE Obtain $Z$ by evaluating the cost function of Problem \ref{p:CaseI_lower_recast} with $\{b_k\}$ replaced with $\{b_k^*\}$.
\STATE \%\% Part II: Obtain the optimal $S$ for solving Problem \ref{p:CaseI_upper}
\STATE With obtained $Z$ in Part I, calculate the optimal $S$ by following (\ref{e:case_I_S}).
\end{algorithmic}
\end{algorithm}

~\\

In terms of the computation complexity of Algorithm \ref{a:case_I}, its mainly comes from the solving of Problem \ref{p:CaseI_lower_recast}, which is essentially a combination of bi-section search of $r$ and a convex optimization problem as shown in Problem \ref{p:CaseI_lower_convex_feasibility}.
As we know, the complexity for solving Problem \ref{p:CaseI_lower_convex_feasibility} is $O((K+1)^3)$. Then the total complexity of Algorithm \ref{a:case_I} is $O(\log_2(\varepsilon_b) (K+1)^3)$, where $\varepsilon_b$ is the tolerance of bi-section search, which is polynomial.

\subsection{Optimization for Case II} \label{s:optimization_CaseII}

In this subsection, the charactered bound of optimality gap for Case II is going to be minimized by adjusting $a$, $\eta$, and $\{b_k | k\in \mathcal{K}\}$. To be exact, we need to minimize the second term of the right-hand side of (\ref{e:Lemma2_ineq}) for given $q^{\max}$, which describes a specific convergence rate.
Considering $q^{\max}$ may be zero or lie between $(0, 1)$ since $q^{\max} = \max \left(\left(1 - \frac{2M \cos (\theta_{\text{th}})\eta a \sum_{k\in \mathcal{K}} h_k b_k}{G}\right), 0\right)$, two discussions can be unfolded:
\subsubsection{When $q^{\max} = 0$}
In this case, according to the definition of $q^{\max}$, there is
\begin{equation} \label{e:q_max_eq_0_cond}
{2M \cos (\theta_{\text{th}}) \eta a \sum_{k\in \mathcal{K}} h_k b_k} \geq G,
\end{equation}
and the second term of the right-hand side of (\ref{e:Lemma2_ineq}) turns out to be
\begin{equation}  \label{e:q_max_eq_0}
\frac{L}{2} a^2 \eta^2 \left(\sum_{k\in \mathcal{K}} 4 h_k^2 b_k^2 + \left(\sum_{k\in \mathcal{K}}h_k b_k \right)^2 + n \sigma^2 \right).
\end{equation}
To minimize the expression in (\ref{e:q_max_eq_0}), by replacing $a \eta$ with its lower bound characterized in (\ref{e:q_max_eq_0_cond}), the problem of minimizing (\ref{e:q_max_eq_0}) can be formulated as
\begin{prob} \label{p:CaseII_convex_q_max_0}
\begin{subequations}
\begin{align}
\mathop{\min} \limits_{\{b_k | k\in \mathcal{K}\}} \quad & \frac{L G^2}{8 M^2 \cos^2(\theta_{\text{th}})} \frac{\left(\sum_{k\in \mathcal{K}} 4 h_k^2 b_k^2 + \left(\sum_{k\in \mathcal{K}} h_k b_k \right)^2 + n \sigma^2 \right)}{\left(\sum_{k\in \mathcal{K}} h_k b_k \right)^2}   \nonumber\\
  \text{s.t.}  \quad & \sum_{k\in \mathcal{K}} h_k b_k >0, \\
\quad &   0 \leq b_k \leq b_k^{\max}, \forall k \in \mathcal{K},	
\end{align}
\end{subequations}
\end{prob}
which can be found to be equivalent with Problem \ref{p:CaseI_lower_recast} exactly. Thus the optimal solution of Problem \ref{p:CaseII_convex_q_max_0} is able to be obtained by following Part I of Algorithm \ref{a:case_I} for solving Problem \ref{p:CaseI_lower_recast}.
Thereafter, the minimal achievable cost function of Problem \ref{p:CaseII_convex_q_max_0} can be expressed as $\frac{LG^2(Z+1)}{8M^2 \cos^2(\theta_{\text{th}})}$.

Collecting the obtained results for the right-hand side of (\ref{e:Lemma2_ineq}) in this case, the first term will be zero irrespective of $T$ because $q^{\max}=0$ and the minimal achievable value for the second term is $\frac{LG^2 (Z+1)}{8M^2 \cos^2(\theta_{\text{th}})}$, which is also larger than zero because $Z>0$.  Hence the bound of optimality gap in this case is not able to achieve zero no matter how to set $T$.

\subsubsection{When $q^{\max} \in (0, 1)$}
In this case, suppose $q^{\max} = s$, where $s \in (0, 1)$, then according to the definition of $q^{\max}$, there is
\begin{equation} \label{e:q_max_ineq_0_cond}
{2M \cos(\theta_{\text{th}}) \eta a \sum_{k\in \mathcal{K}} h_k b_k} = G \left(1 - s\right),
\end{equation}
and the second term of the right-hand side of (\ref{e:Lemma2_ineq}) turns out to be
\begin{equation}  \label{e:q_max_bigg_0}
\frac{a \eta L G}{4 M \cos(\theta_{\text{th}}) \sum_{k\in \mathcal{K}} h_k b_k}  \left(\sum_{k\in \mathcal{K}} 4 h_k^2 b_k^2 + \left(\sum_{k\in \mathcal{K}}h_k b_k \right)^2 + n \sigma^2 \right).
\end{equation}

Replace $a \eta$ according to (\ref{e:q_max_ineq_0_cond}), then the problem of minimizing (\ref{e:q_max_bigg_0}) can be formulated as
\begin{prob} \label{p:CaseII_convex_q_max_bigg_0}
\begin{subequations}
\begin{align}
C_2^{\text{II}}(s) = & \notag \\
 \mathop{\min} \limits_{\{b_k | k\in \mathcal{K}\}} \quad &  \frac{L G^2 \left(1 - s\right) \left(\sum_{k\in \mathcal{K}} 4 h_k^2 b_k^2 + \left(\sum_{k\in \mathcal{K}}h_k b_k \right)^2 + n \sigma^2 \right) }{8 M^2 \cos^2(\theta_{\text{th}}) \left(\sum_{k\in \mathcal{K}} h_k b_k \right)^2 }
\nonumber\\
  \text{s.t.}  \quad & \sum_{k\in \mathcal{K}} h_k b_k >0, \\
\quad &   0 \leq b_k \leq b_k^{\max}, \forall k \in \mathcal{K}.
\end{align}
\end{subequations}
\end{prob}

For Problem \ref{p:CaseII_convex_q_max_bigg_0}, it is also equivalent with Problem \ref{p:CaseI_lower_recast} and can be solved optimally by following Part I of Algorithm \ref{a:case_I} for solving Problem \ref{p:CaseI_lower_recast}.
Moreover, recalling that the minimal cost function of Problem \ref{p:CaseI_lower_recast} is defined as $Z$, the minimal cost function of Problem \ref{p:CaseII_convex_q_max_bigg_0} can be written as $C_2^{\text{II}}(s) = \frac{(Z+1)LG^2(1-s)}{8 M^2 \cos^2(\theta_{\text{th}})}$, which is a linear decreasing function with $s$. With these results, we have the following two observations:
\begin{itemize}
\item Achieving $\varepsilon$-small training loss at linear rate: For any predefined $\varepsilon>0$, by setting $s =\left(1 - \frac{8 M^2 \cos^2(\theta_{\text{th}}) \varepsilon}{(Z+1)LG^2}\right)$, which is surely no larger than 1, the second term of (\ref{e:Lemma2_ineq})'s right-hand side expression, which is taken as the bias item, can be guaranteed to be no larger than $\varepsilon$. Then if $\varepsilon$ is an infinite small value close to zero, so will the second term of the right-hand side of (\ref{e:Lemma2_ineq}). On the other hand, as $s$ is still less than 1, the first term of the right-hand side of (\ref{e:Lemma2_ineq}) can approach to zero at linear rate as $T$ goes into infinity.
\item Tradeoff between the convergence rate $q^{\max}$ and the bias item $\varepsilon$: To achieve a lower $\varepsilon$, $s$ has to be as close to 1 as possible, which will lead to a higher $q^{\max}$, and thus a lower convergence rate for the first term of (\ref{e:Lemma2_ineq})'s right-hand side expression. Hence there is a tradeoff between the convergence rate $q^{\max}$ and the bias item $\varepsilon$. The preference on either one can be realized by adjusting the parameter $s$.
\end{itemize}

At the end of this subsection, three points can be summarized in the following remark  for Case II.
\begin{remark} \label{r:CaseII}
\begin{itemize}
\item By analyzing the cases that $q^{\max}=0$ and $q^{\max} \in (0, 1)$, which can achieve non-zero optimality gap bound and any $\varepsilon$-small optimality gap bound at a linear rate, respectively. It would be a better choice to select $q^{\max}$ to be within  $(0,1)$. With such a selection, also recalling that $F(\bm{w})$ is strongly convex, we can claim that the global optimal solution can be achieved with a tolerance of $\varepsilon$ in Case II.
\item By selecting a $q^{\max} \in (0, 1)$, a tradeoff exists between the $q^{\max}$ and the bias term $\varepsilon$.
\item Compared with the results for Case I as shown in Remark \ref{r:CaseI}, which merely assumes smoothness on loss function and achieves a sub-linear convergence rate, the additional condition of strong convexity in Case II brings about the benefit of reaching a linear convergence rate, which is faster than sub-linear convergence rate.
\end{itemize}
\end{remark}

\section{Experiments} \label{s:exp}
In this section, numerical results are presented to validate the performance of our proposed methods under two investigated cases. In default, the number of devices {$K=20$}. 
The wireless channels $h_k$ for $k\in \mathcal{K}$ are subjected to i.i.d. Rayleigh distribution with mean being {$10^{-5}$}, and the variance of noise {$\sigma^2=10^{-7}$} \footnote{This mean value of $h_k$ is obtained with free-space attenuation over a distance {300} meters at a carrier frequency {3.5GHz} in composition with a Rayleigh distribution with mean being 1. The $\sigma^2$ is calculated by multiplying the power spectrum density {-140 dBW/Hz} with a bandwidth of {10MHz} \cite{Iot_Li}.}.
$b_k^{\max}$ are all set to be $\sqrt{5}$ for $k\in \mathcal{K}$.
{$\theta_{\text{th}}$ is assumed to be $\frac{\pi}{3}$.}
For case I, a handwritten 10-digit recognition task based on MNIST dataset is performed. The classifier like the one in \cite{Cui_JSAC} is selected, which has three fully connected layers, one ReLU activation layer, and one SoftMax output layer. The loss function in this case is smooth but non-convex.
The learning rate $\eta^{(t)}= 1/t^{0.75}$ and batch size is set as 50.
For case II, a ridge regression model like \cite{Cui_JSAC} will be trained, whose loss function is shown to be strongly convex.
The associated learning rate $\eta^{(t)} = \eta=0.01$.
Two benchmark methods are compared \cite{Cui_JSAC,Angle_RIS_TWC} with our proposed method, which can achieve best convergence performance by far without and with unifying the local gradient to be a norm-1 vector before amplification, and are abbreviated as ``Benchmark I'' and ``Benchmark II'', respectively.
When the comparison is made between our proposed method and the benchmark methods, test accuracy (in terms of correct prediction rate)  and loss value (the value of loss function) are taken into account for the training task in Case I and Case II, respectively.



\subsection{Performance Comparison}


\begin{figure}
	\centering
	\subfigure[The benefit of optimizing $a$ and $\{b_k\}$.]{
		\begin{minipage}{0.45\linewidth}
			\centering
			\includegraphics[width=0.95\linewidth]{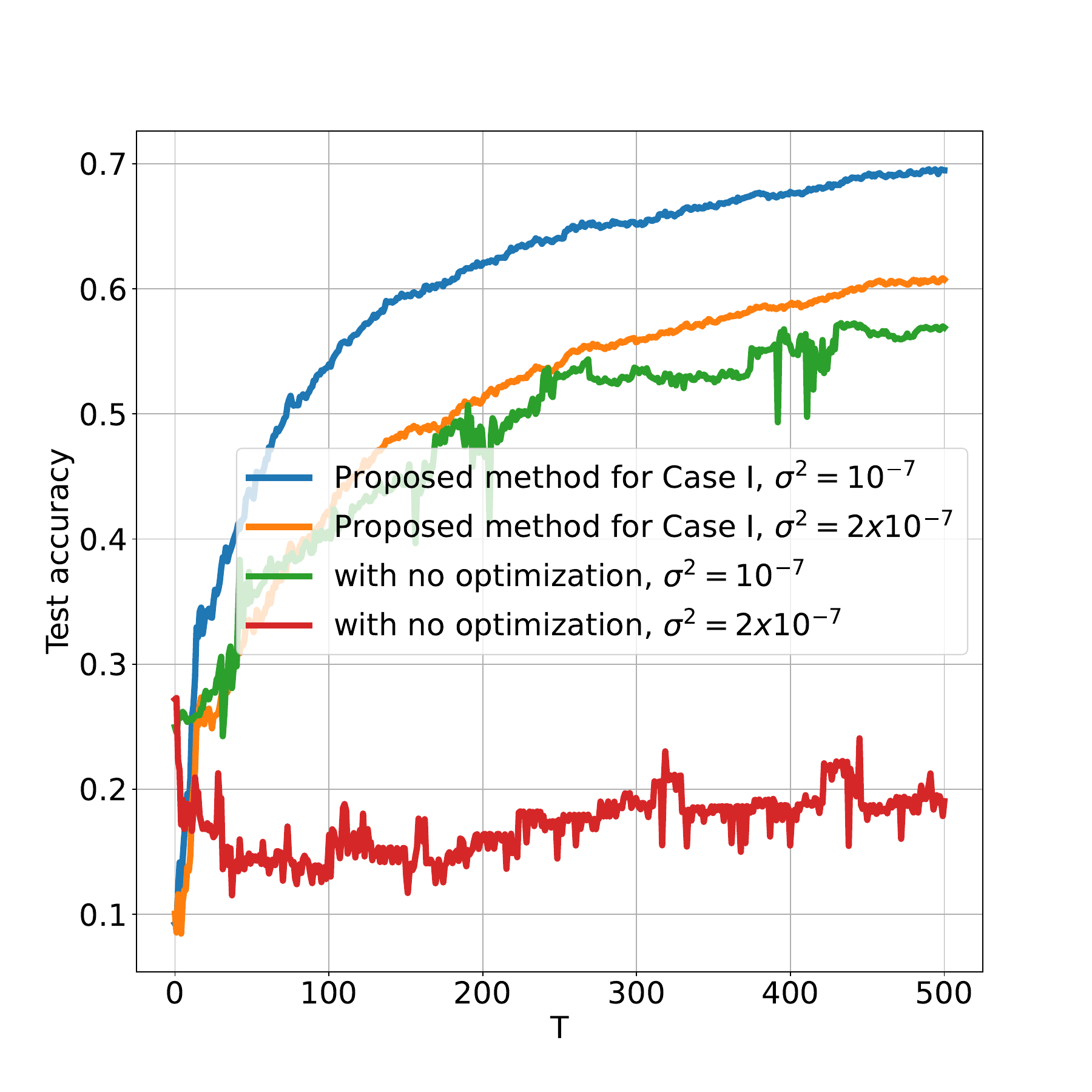}
			\label{class:opt}
		\end{minipage}
	}
	\subfigure[Comparison with benchmarks.]{
		\begin{minipage}{0.45\linewidth}
			\centering
			\includegraphics[width=0.95\linewidth]{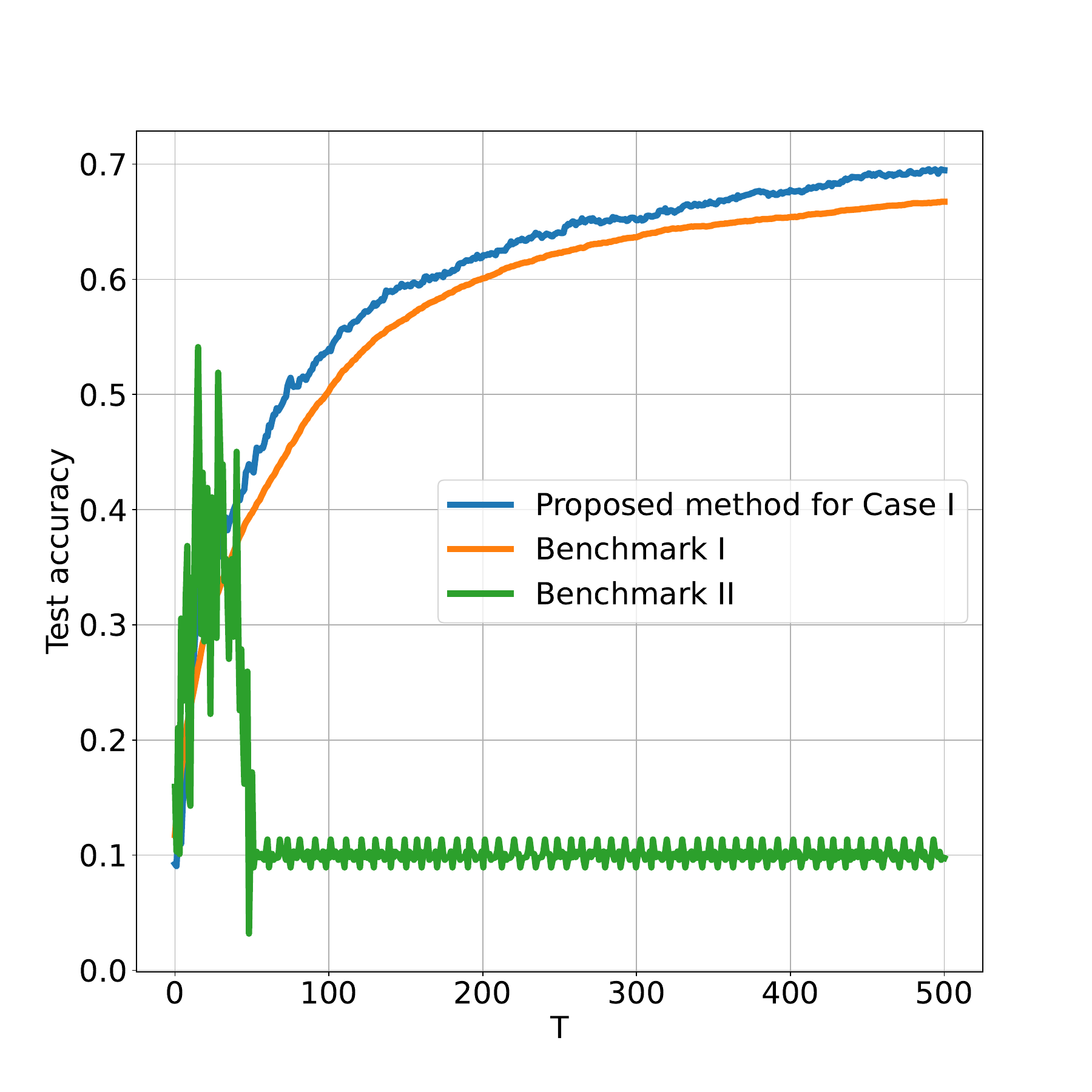}
			\label{class:acc}
		\end{minipage}
	}
	\caption{Performance comparison for Case I.}
	\label{air-re}
\end{figure}

\begin{figure}
	\centering
	\subfigure[The benefit of optimizing $a$ and $\{b_k\}$.]{
		\begin{minipage}{0.45\linewidth}
			\centering
			\includegraphics[width=0.95\linewidth]{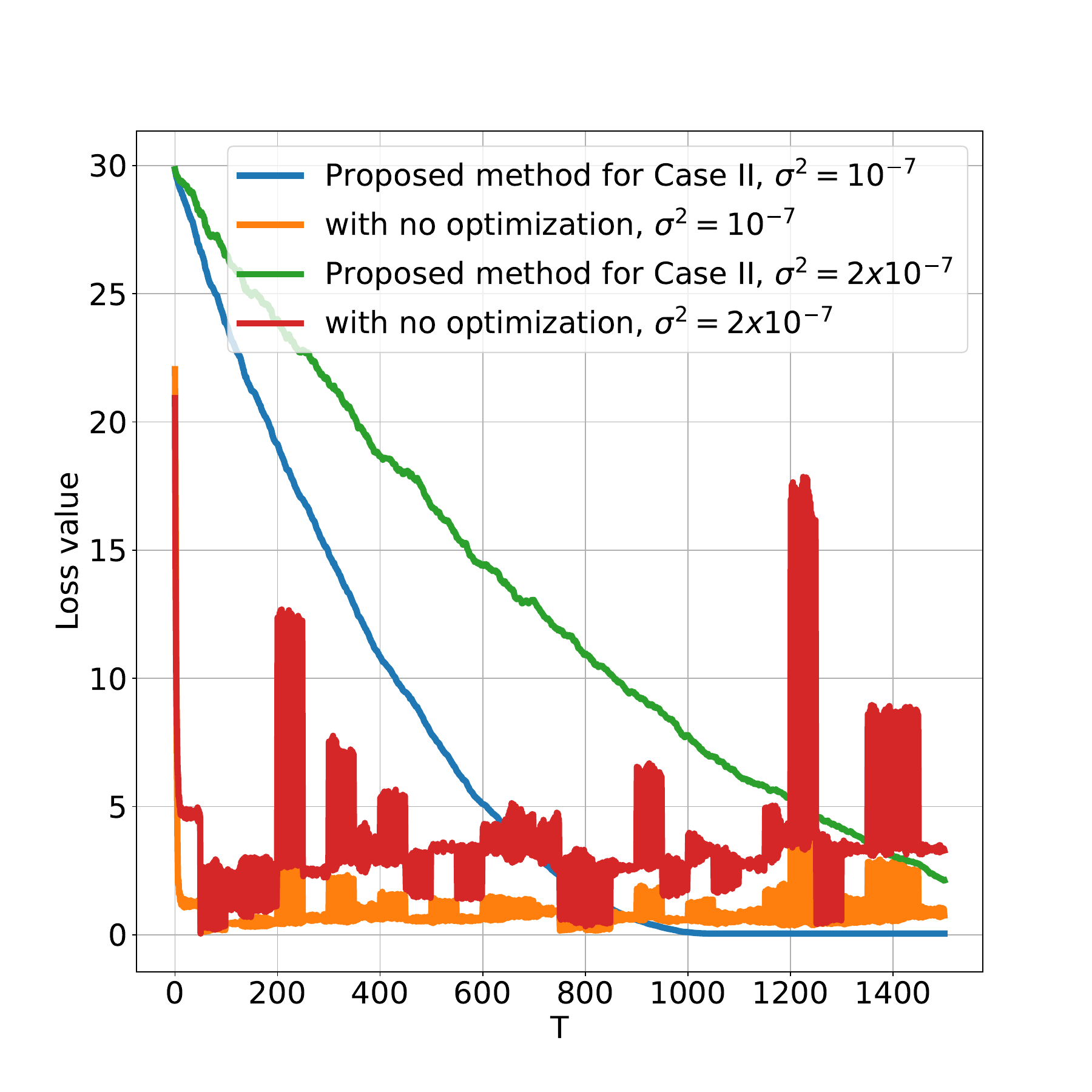}
			\label{linear:opt}
		\end{minipage}
	}
	\subfigure[Comparison with benmarks.]{
		\begin{minipage}{0.45\linewidth}
			\centering
			\includegraphics[width=0.95\linewidth]{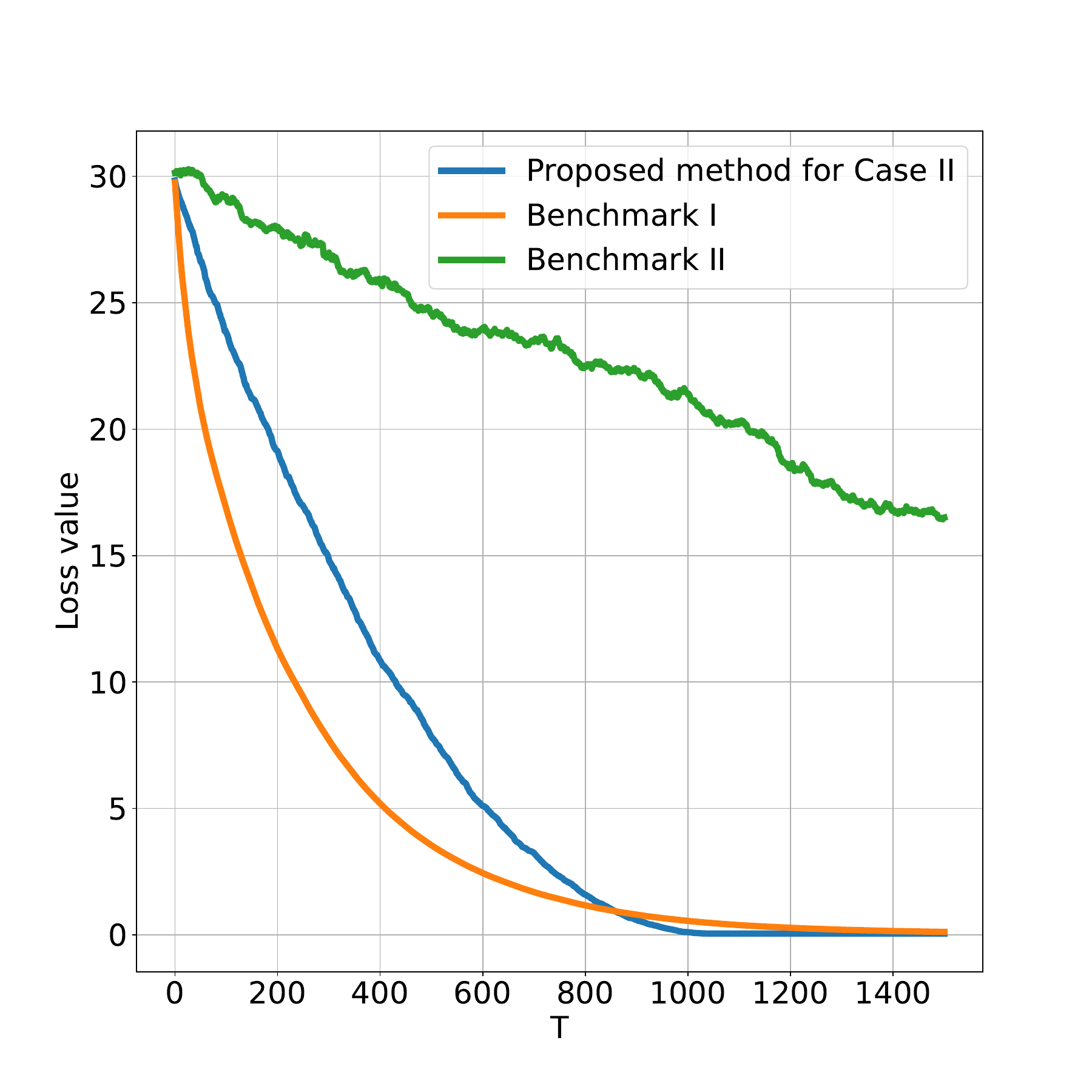}
			\label{linear:err}
		\end{minipage}
	}
	\caption{Performance comparison for Case II.}
	\label{air-lin}
\end{figure}

Fig. \ref{air-re} shows performance comparison results for our proposed method under Case I.
To be exactly, Fig. \ref{class:opt} plots the test accuracy of our proposed method as $T$ grows. As a comparison, another method which adopts our proposed aggregation method but does not optimize $a$ and $\{b_k\}$ by simply setting $b_k= b_k^{\max}$ and selecting the $a$ such that $a \sum_{k}b_k$ to be equal with the one in our proposed method. It can be observed that our proposed method can really speedup convergence in contrast to the compared one, which verifies the benefit of optimizing $a$ and $\{b_k\}$.
In Fig.  \ref{class:acc}, our proposed is compared with two benchmarks methods in \cite{Cui_JSAC} and \cite{Angle_RIS_TWC}. It can be also seen that our proposed method can always achieve higher test accuracy as $T$ grows, which proves the advantage of our proposed one over existing methods. Similar results can be also obtained for Case II, as shown in Fig. \ref{linear:opt}, and is omitted here due to limited space.

\begin{figure}
	\centering
	\subfigure[Extra benefit of strong convexity]{
		\begin{minipage}{0.45\linewidth}
			\centering
			\includegraphics[width=0.95\linewidth]{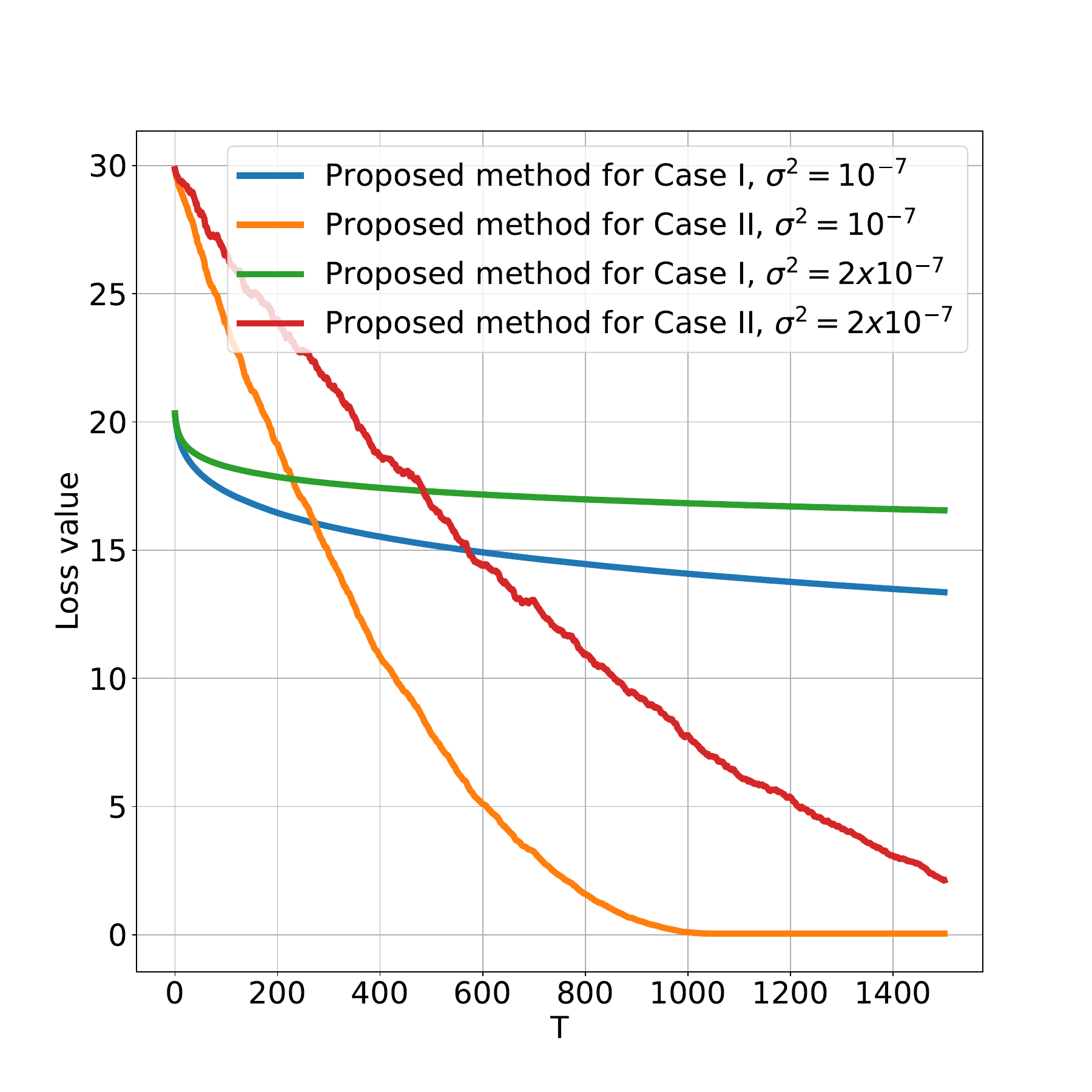}
			\label{f:more_inf}
		\end{minipage}
	}
	\subfigure[Tradeoff between $\varepsilon$ and $q^{\max}$]{
		\begin{minipage}{0.45\linewidth}
			\centering
			\includegraphics[width=0.95\linewidth]{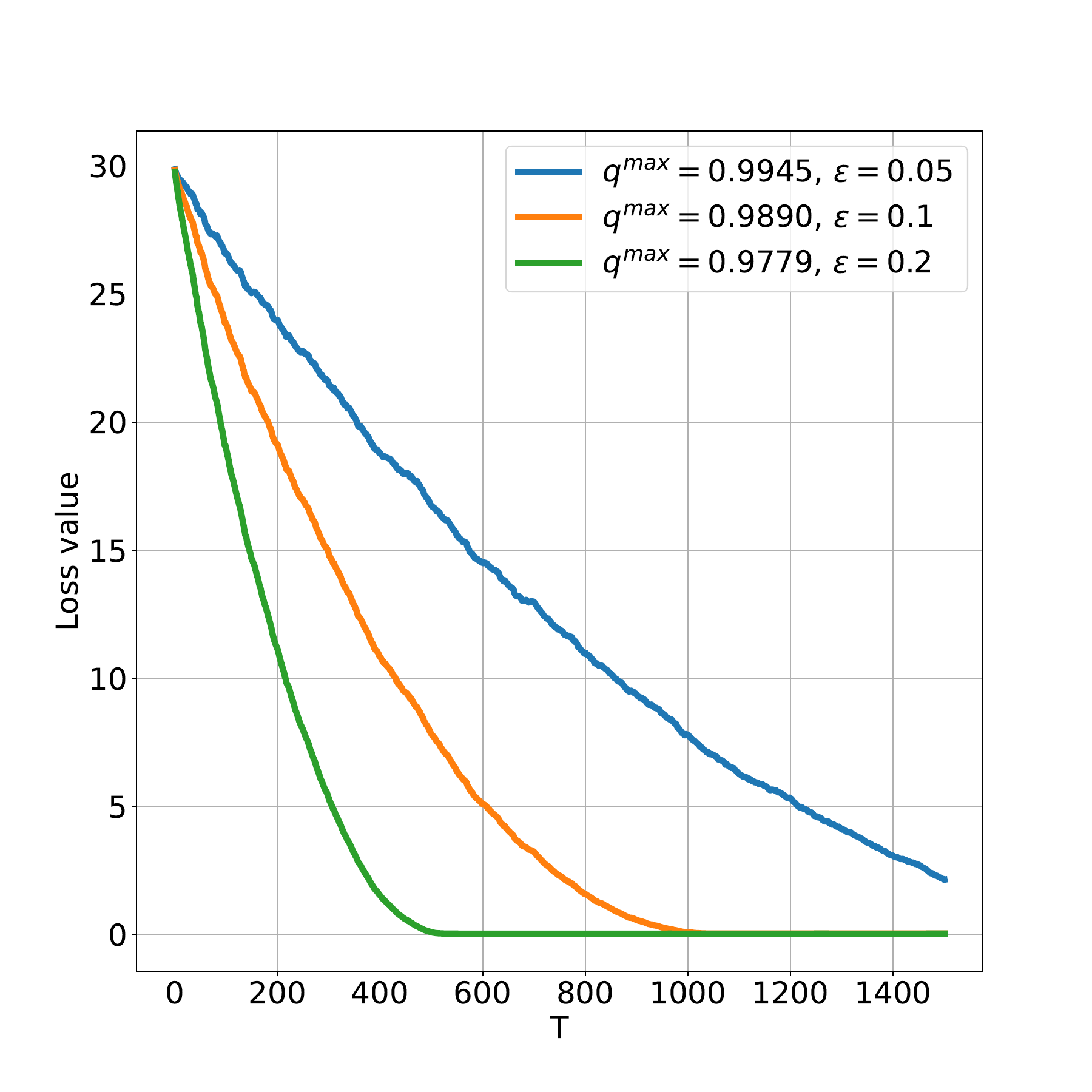}
			\label{f:trade}
		\end{minipage}
	}
	\caption{More results of our proposed methods.}
	\label{lin-abal}
\end{figure}

\subsection{Performance Analysis}
In Fig. \ref{f:more_inf}, both of our proposed method for Case I and Case II are utilized to run the ridge regression model, whose loss function is not only smooth but also strongly-convex and minimal training loss is achievable by our proposed method for either case.
It can be found that our proposed method for Case II can converge at much faster speed than the one for Case I, which discloses the benefit of exploiting the strong convexity of the loss function, when it is not only smooth but also strongly-convex.
In Fig. \ref{f:trade}, the loss value in Case II is plotted versus $T$ when $q^{\max}$ is 0.9945, 0.9890, and 0.9779, which corresponds to a $\varepsilon$ value of 0.05, 0.1, 0.2, respectively.
It can be observed that when $q^{\max}$ is larger, which implies a smaller $\varepsilon$ and thus lower gap to the minimal training loss, the associated loss value can converge at a faster speed. This verifies the characterized tradeoff in Remark \ref{r:CaseII}.

\section{Conclusion} \label{s:conclu}
In this paper, we have proposed a new aggregation method for an over-the-air computation aided FL system, which normalizes the local gradient at every mobile device before amplifying it.
With our proposed method,
stationary point is achieved at a sub-linear rate when the loss function is smooth only, and global optimal solution can be obtained at a linear rate with $\varepsilon$-tolerance when the loss function is not only smooth but also strongly convex, followed by the disclosure of a tradeoff between the convergence rate and the $\varepsilon$-tolerance.
System parameters are also optimized under the above two considered cases so as to further speedup convergence, both of which correspond to a non-convex problem.
Optimal solution with polynomial complexity for the formulated problems are offered.

\begin{appendices}

\section{Proof of Lemma \ref{lem:CaseI}} \label{app:lemma1}

With Assumption \ref{a:smooth} and according to \cite{Nesterov}, there is
\begin{subequations} \label{e:gap_F_unexpected}
\begin{align}
& F(\bm{w}^{t+1}) - F(\bm{w}^t) \notag \\
\leq &  \nabla F(\bm{w}^t)^T (\bm{w}^{(t+1)} - \bm{w}^{(t)}) + \frac{L}{2} \left \lVert \bm{w}^{(t+1)} - \bm{w}^{(t)} \right \rVert^2 \\
\leq & -\nabla F(\bm{w}^{(t)})^T  \eta^{(t)} a \left( \sum_{k\in \mathcal{K}} h_k b_k \cdot {\bm{g}_k^{(t)}}/{\lVert \bm{g}_k^{(t)} \rVert} + \bm{z}^{(t)} \right)
+ \frac{1}{2} L a^2 \left(\eta^{(t)}\right)^2 \left \lVert \sum_{k\in \mathcal{K}} h_k b_k \cdot {\bm{g}_k^{(t)}}/{\lVert \bm{g}_k^{(t)} \rVert} + \bm{z}^{(t)} \right \rVert^2
\end{align}
\end{subequations}
Take expectation on both sides of the equation (\ref{e:gap_F_unexpected}), there is the expression of (\ref{e:gap_F_expect}).
In (\ref{e:gap_F_expect}),
\begin{itemize}
\item the inequality in (\ref{e:gap_F_expect_first}) holds because the angle between $\bm{g}_k^{(t)}$ and the vector $\nabla F(\bm{w}^{(t)})$, i.e., the random variable $\theta_k$, is also the angle between ${\bm{g}_k^{(t)}}/{\lVert \bm{g}_k^{(t)} \rVert}$ and the vector ${\nabla F(\bm{w}^{(t)}) }/{\lVert \nabla F(\bm{w}^{(t)}) \rVert }$, and the fact $\mathbb{E}\{\bm{z}^{(t)}\} = \bm{0}$;

\item the  inequality in (\ref{e:gap_F_expect_second}) comes from the fact $|\theta_{k}|< \theta_{\text{th}}$ $\left(\theta_{\text{th}} < \pi/2\right)$ according to Assumption \ref{a:bias}, the fact that $\bm{z}^t$ is independent from any random variable related to $g_k^{(t)}$, and the fact that $\mathbb{E}\{\bm{z}^{(t)}\} = \bm{0}$;

\item the inequality in (\ref{e:gap_F_expect_third}) is established since $g_k^{(t)}$ and $g_{k'}^{(t)}$ are independent from each other according to Assumption \ref{a:independent}, the random variable $\left( {\bm{g}_k^{(t)}}/{\lVert \bm{g}_k^{(t)} \rVert} - \mathbb{E} \left\{{\bm{g}_k^{(t)}}/{\lVert \bm{g}_k^{(t)} \rVert} \right\} \right)$ has zero mean, and the fact that
$\left \lVert \mathbb{E} \left\{ {\bm{g}_k^{(t)}}/{\lVert \bm{g}_k^{(t)} \rVert} \right\} \right \rVert \leq \mathbb{E} \left\{  \left \lVert {\bm{g}_k^{(t)}}/{\lVert \bm{g}_k^{(t)} \rVert}  \right \rVert \right\} = 1$,
due to the convexity of norm function $\lVert \cdot \rVert$, which further implies \\
$\left \lVert \left( {\bm{g}_k^{(t)}}/{\lVert \bm{g}_k^{(t)} \rVert} - \mathbb{E} \left\{{\bm{g}_k^{(t)}}/{\lVert \bm{g}_k^{(t)} \rVert} \right\} \right) \right \rVert^2 \leq  \left \lVert {\bm{g}_k^{(t)}}/{\lVert \bm{g}_k^{(t)} \rVert} - \left({-\bm{g}_k^{(t)}}/{\lVert \bm{g}_k^{(t)} \rVert} \right) \right \rVert^2$;
\item the inequality in (\ref{e:gap_F_expect_fourth}) is supported by the facts that $\mathbb{E} \left\{ \cos (\theta_k)\right\} \leq 1$ and random variable $\left( {\bm{g}_k^{(t)}}/{\lVert \bm{g}_k^{(t)} \rVert} - \mathbb{E} \left\{{\bm{g}_k^{(t)}}/{\lVert \bm{g}_k^{(t)} \rVert} \right\} \right)$ has zero mean.
\end{itemize}
\begin{subequations} \label{e:gap_F_expect}
\begin{align}
& \mathbb{E} \left\{F(\bm{w}^{(t+1)}) - F(\bm{w}^{(t)}) \right \}  \notag \\
\leq &  - \eta^{(t)} a \left( \sum_{k\in \mathcal{K}} h_k b_k \right)  \left(\nabla F(\bm{w}^{(t)})^T  \frac{ \nabla F(\bm{w}^{(t)}) \mathbb{E} \left\{\cos(\theta_k) \right\}}{\lVert  \nabla F(\bm{w}) \rVert}\right)  \notag \\
& + \frac{1}{2} L a^2 \left(\eta^{(t)} \right)^2  \times  \mathbb{E} \Bigg\{ \Bigg \lVert \sum_{k\in \mathcal{K}} h_k b_k  \left( \frac{\bm{g}_k^{(t)}}{\lVert \bm{g}_k^{(t)} \rVert} -  \mathbb{E} \left\{ \frac{\bm{g}_k^{(t)}}{\lVert \bm{g}_k^{(t)} \rVert} \right\} \right)
 +  \sum_{k\in \mathcal{K}} h_k b_k \mathbb{E} \left\{ \frac{\bm{g}_k^{(t)}}{\lVert \bm{g}_k^{(t)} \rVert} \right\}  + \bm{z}^{(t)} \Bigg \rVert^2 \Bigg\} \label{e:gap_F_expect_first}\\
\leq &  - \eta^{(t)} a \left( \sum_{k\in \mathcal{K}} h_k b_k \right)  \left(\nabla F(\bm{w}^{(t)})^T   \frac{ \nabla F(\bm{w}^{(t)}) }{\lVert \nabla F(\bm{w}^{(t)}) \rVert } \cos (\theta_{\text{th}}) \right) \notag \\
& + \frac{1}{2} L a^2 \left(\eta^{(t)} \right)^2 \Bigg( \mathbb{E} \left\{ \left \lVert \sum_{k\in \mathcal{K}} h_k b_k  \left( \frac{\bm{g}_k^{(t)}}{\lVert \bm{g}_k^{(t)} \rVert} -  \mathbb{E} \left\{ \frac{\bm{g}_k^{(t)}}{\lVert \bm{g}_k^{(t)} \rVert} \right\} \right) \right \rVert ^2 \right\}
 +  \mathbb{E} \left \{ \left \lVert \sum_{k\in \mathcal{K}} h_k b_k \mathbb{E} \left\{ \frac{\bm{g}_k^{(t)}}{\lVert \bm{g}_k^{(t)} \rVert} \right\} \right \rVert^2 \right \}   + \mathbb{E} \left \{ \left \lVert \bm{z}^{(t)} \right \rVert^2 \right\}  \notag \\
& + 2 \mathbb{E} \Bigg \{ \left(\sum_{k\in \mathcal{K}} h_k b_k \left( \frac{\bm{g}_k^{(t)}}{\lVert \bm{g}_k^{(t)} \rVert} -  \mathbb{E} \left\{ \frac{\bm{g}_k^{(t)}}{\lVert \bm{g}_k^{(t)} \rVert} \right\} \right)\right)^T  \cdot
 \left(\sum_{k\in \mathcal{K}} h_k b_k \mathbb{E} \left\{ \frac{\bm{g}_k^{(t)}}{\lVert \bm{g}_k^{(t)} \rVert} \right\} \right)\Bigg \} \Bigg) \label{e:gap_F_expect_second}\\
\leq &   - \eta^{(t)} a \left( \sum_{k\in \mathcal{K}} h_k b_k \right)  \lVert \nabla F(\bm{w}^{(t)}) \rVert \cos (\theta_{\text{th}})
+ \frac{1}{2} L a^2 \left(\eta^{(t)}\right)^2 \times \Bigg( \sum_{k\in \mathcal{K}} \mathbb{E} \left\{ \left \lVert  h_k b_k  \left( \frac{\bm{g}_k^{(t)}}{\lVert \bm{g}_k^{(t)} \rVert} - \frac{-\bm{g}_k^{(t)}}{\lVert \bm{g}_k^{(t)} \rVert}  \right) \right \rVert ^2 \right\}  \notag \\
& +  \mathbb{E} \left \{ \left \lVert \sum_{k\in \mathcal{K}} h_k b_k \frac{ \nabla F(\bm{w}^{(t)}) \mathbb{E} \left\{\cos(\theta_k) \right\}}{\lVert  \nabla F(\bm{w}) \rVert} \right \rVert^2 \right \} + n  \sigma^2
+  2 \sum_{k\in \mathcal{K}} h_k^2 b_k^2 \mathbb{E} \left\{\left( \frac{\bm{g}_k^{(t)}}{\lVert \bm{g}_k^{(t)} \rVert} -  \mathbb{E} \left\{ \frac{\bm{g}_k^{(t)}}{\lVert \bm{g}_k^{(t)} \rVert} \right\} \right)^T \mathbb{E} \left\{ \frac{\bm{g}_k^{(t)}}{\lVert \bm{g}_k^{(t)} \rVert} \right\}\right\} \Bigg)  \label{e:gap_F_expect_third} \\
\leq &   - \eta^{(t)} a \left( \sum_{k\in \mathcal{K}} h_k b_k \right)  \lVert \nabla F(\bm{w}^{(t)}) \rVert \cos (\theta_{\text{th}})   \notag \\
& + \frac{1}{2} L a^2 \left(\eta^{(t)}\right)^2 \times \Bigg( \sum_{k\in \mathcal{K}} \mathbb{E} \left\{ \left \lVert 2 h_k b_k  \frac{ \nabla F(\bm{w}^{(t)}) }{\lVert  \nabla F(\bm{w}) \rVert}  \right \rVert ^2 \right\}
+  \mathbb{E} \left \{ \left \lVert \sum_{k\in \mathcal{K}} h_k b_k \frac{ \nabla F(\bm{w}^{(t)}) }{\lVert  \nabla F(\bm{w}) \rVert} \right \rVert^2 \right \}   + n \sigma^2 \Bigg)  \label{e:gap_F_expect_fourth} \\
\leq &   - \eta^{(t)} a \left( \sum_{k\in \mathcal{K}} h_k b_k \right)  \lVert \nabla F(\bm{w}^{(t)}) \rVert \cos (\theta_{\text{th}})
+ \frac{1}{2} L a^2 \left(\eta^{(t)}\right)^2 \left(\sum_{k\in \mathcal{K}} 4 h_k^2 b_k^2  + \left(\sum_{k\in \mathcal{K}} h_k b_k \right)^2 + n \sigma^2  \right). \label{e:gap_F_expect_fifth}
\end{align}
\end{subequations}

Summing up (\ref{e:gap_F_expect}) together for $t=1$, $t=2$, ..., $t=T$, there is
\begin{subequations}
\begin{align}
& \mathbb{E} \left\{ F(\bm{w}^{(T+1)}) - F(\bm{w}^{(1)}) \right \} \notag \\
\leq &  - \sum_{t=1}^T \eta^{(t)} a \left( \sum_{k\in \mathcal{K}} h_k b_k \right)  \lVert \nabla F(\bm{w}^{(t)}) \rVert \cos (\theta_{\text{th}}) \\ \notag
& + \sum_{t=1}^T \frac{1}{2} L a^2 \left(\eta^{(t)}\right)^2 \left(\sum_{k\in \mathcal{K}} 4 h_k^2 b_k^2  + \left(\sum_{k\in \mathcal{K}} h_k b_k \right)^2 + n \sigma^2 \right) \\
\leq &  -  \eta^{(T)} a \left( \sum_{k\in \mathcal{K}} h_k b_k \right)  \sum_{t=1}^T  \lVert \nabla F(\bm{w}^{(t)}) \rVert \cos (\theta_{\text{th}}) \\ \notag
& +  \sum_{t=1}^T \frac{1}{2} L a^2 \left(\eta^{(t)} \right)^2 \left(\sum_{k\in \mathcal{K}} 4 h_k^2 b_k^2  + \left(\sum_{k\in \mathcal{K}} h_k b_k \right)^2 + n \sigma^2 \right)
\end{align}
\end{subequations}
where the second inequality holds because $\eta^{(T)} \leq \eta^{(t)}$ for $ t \in [1, T]$ when $\eta^{(t)}$ is set as $1/t^p$ with $ 1/2 < p < 1$. Then there is
\begin{subequations} \label{e:gap_F_through_ineq}
\begin{align}
& \mathop{\min} \limits_{t \in [0,T]}  \lVert \nabla F(\bm{w}^{(t)}) \rVert \notag \\
\leq &  \frac{1}{T} \left(\sum_{t=1}^{T} \lVert \nabla F(\bm{w}^{(t)}) \rVert  \right) \\
\leq & \frac{1}{T \cos (\theta_{\text{th}}) } \frac{\mathbb{E} \left \{F(\bm{w}^{(1)}) - F(\bm{w}^{(T+1)})\right \}}{\eta^{(T)} a \sum_{k\in \mathcal{K}} h_k b_k } \notag \\
& + \frac{1}{T \cos (\theta_{\text{th}})} \frac{1}{\eta^{(T)} \sum_{k\in \mathcal{K}} h_k b_k} \frac{a L}{2} \times \left(\sum_{t=1}^T \left(\eta^{(t)}\right)^2\right) \times \left( \sum_{k\in \mathcal{K}} 4 h_k^2 b_k^2 + \left(\sum_{k\in \mathcal{K}} h_k b_k \right)^2 + n \sigma^2 \right) \label{e:gap_F_through_ineq_second} \\
\leq & \frac{\mathbb{E}\left \{F(\bm{w}^{(1)}) - F(\bm{w}^{(T+1)})\right\}}{ T^{1-p} \cdot a \cos (\theta_{\text{th}}) \sum_{k\in \mathcal{K}} h_k b_k }  \notag \\
&
+ \frac{1}{T^{1-p}} \left(\sum_{t=1}^T \frac{1}{t^{2p}} \right)  \times   \frac{a L }{2 \cos (\theta_{\text{th}})  \sum_{k\in \mathcal{K}} h_k b_k} \times \left( \sum_{k\in \mathcal{K}} 4 h_k^2 b_k^2 + \left(\sum_{k\in \mathcal{K}} h_k b_k \right)^2 + n \sigma^2 \right) \label{e:gap_F_through_ineq_third} \\
\leq & \frac{\mathbb{E}\left \{F(\bm{w}^{(1)}) - F(\bm{w}^{(T+1)})\right\}}{ T^{1-p} \cdot a \cos (\theta_{\text{th}}) \sum_{k\in \mathcal{K}} h_k b_k }   \notag \\
& + \frac{1}{T^{1-p}} \left(1 + \int_1^{T} x^{-2p} dx \right)  \times  \frac{a L \left( \sum_{k\in \mathcal{K}} 4 h_k^2 b_k^2 + \left(\sum_{k\in \mathcal{K}} h_k b_k \right)^2 + n \sigma^2 \right) }{2 \cos (\theta_{\text{th}})  \sum_{k\in \mathcal{K}} h_k b_k}  \label{e:gap_F_through_ineq_fourth}.
\end{align}
\end{subequations}

Look into right-hand side of the inequality (\ref{e:gap_F_through_ineq_fourth}),
the $\frac{1}{T^{1-p}}$ in the first term and second term will converge to zero sub-linearly as $T$ goes to infinity for $p<1$, and the expression $\left(1 + \int_1^{T} x^{-2p} dx \right)$ in the second term is upper bounded by $\frac{2p}{2p -1}$ as $2p>1$.
To sum up, the right-hand side of the inequality (\ref{e:gap_F_through_ineq_fourth}) will converge to zero as $T$ goes to infinity for $1/2 < p <1$.

This completes the proof.

\section{Proof of Lemma \ref{lem:strong_convex}} \label{app:lemma2}

With Assumption \ref{a:smooth}, according to \cite{Nesterov}, there is
\begin{equation}
\begin{array}{ll} \label{e:F_w_convert}
& F(\bm{w}^{(t+1)}) - F(\bm{w}^*)  \\
\leq & \nabla F(\bm{w}^*) (\bm{w}^{t+1} - \bm{w}^*) + \frac{L}{2} \left \lVert \bm{w}^* - \bm{w}^{(t+1)} \right \rVert^2 \\
=  & \frac{L}{2} \left \lVert \bm{w}^* - \bm{w}^{(t+1)} \right \rVert^2
\end{array}
\end{equation}
which holds since $\nabla F(\bm{w}^*) = 0$.

Then we investigate how $\lVert \bm{w}^{(T)} - \bm{w}^* \rVert^2$ varies with $T$. In the first step, the inequality as shown in (\ref{e:ineq_smooth_strong}) can be found,
\begin{subequations}  \label{e:ineq_smooth_strong}
\begin{align}
& \mathbb{E} \left\{ \lVert \bm{w}^{(t+1)} - \bm{w}^* \rVert^2 \right\} \notag \\
= & \mathbb{E} \left\{ \left \lVert \bm{w}^{(t)} - \eta^{(t)} a \left(\sum_{k\in \mathcal{K}} h_k b_k \frac{\bm{g}_k^{(t)}}{\lVert \bm{g}_k^{(t)} \rVert} + \bm{z}_k \right) - \bm{w}^{*} \right \rVert^2 \right \} \label{e:ineq_smooth_strong_first} \\
= & \mathbb{E} \left \{\lVert \bm{w}^{(t)} - \bm{w}^* \rVert^2 - 2 \left(\bm{w}^{(t)} - \bm{w}^* \right)^T \eta^{(t)} a \left(\sum_{k\in \mathcal{K}} h_k b_k \frac{\bm{g}_k^{(t)}}{\lVert \bm{g}_k^{(t)} \rVert} \right) \right\}   \notag \\
& + \left(\eta^{(t)}\right)^2 a^2  \mathbb{E} \Bigg\{ \Bigg \lVert \sum_{k\in \mathcal{K}} h_k b_k \left(\frac{\bm{g}_k^{(t)}}{\lVert \bm{g}_k^t \rVert}  - \mathbb{E} \left \{\frac{\bm{g}_k^{(t)}}{\lVert \bm{g}_k^{(t)} \rVert} \right \} \right)
+ \sum_{k\in \mathcal{K}} h_k b_k \mathbb{E} \left\{ \frac{\bm{g}_k^{(t)}}{\lVert \bm{g}_k^{(t)} \rVert} \right \}  + \bm{z}_k^{(t)} \Bigg \rVert^2 \Bigg \} \label{e:ineq_smooth_strong_second} \\
\leq & \mathbb{E} \Bigg \{\lVert \bm{w}^{(t)} - \bm{w}^* \rVert^2 - 2 \left(\bm{w}^{(t)} - \bm{w}^* \right)^T \eta^{(t)} a
 \times \left(\sum_{k\in \mathcal{K}} h_k b_k \frac{ \nabla F(\bm{w}^{(t)} ) \mathbb{E} \left\{ \cos (\theta_{k}) \right \} - \nabla F(\bm{w}^*) \mathbb{E} \left\{ \cos (\theta_{k}) \right \}}{ \lVert \nabla F(\bm{w}^{(t)} \rVert} \right) \Bigg \}  \notag \\
& + \left(\eta^{(t)}\right)^2 a^2 \left(\sum_{k\in \mathcal{K}} 4 h_k^2 b_k^2 + \left(\sum_{k\in \mathcal{K}} h_k b_k \right)^2 + n \sigma^2 \right) \label{e:ineq_smooth_strong_third} \\
\leq & \left( 1- \frac{2 M \eta^{(t)} a \sum_{k\in \mathcal{K}} h_k b_k \mathbb{E} \left\{ \cos (\theta_{k}) \right \}}{\lVert \nabla F(\bm{w}^{(t)}) \rVert} \right) \mathbb{E} \left \{\lVert \bm{w}^{(t)} - \bm{w}^* \rVert^2 \right \}
+ \left(\eta^{(t)}\right)^2 a^2 \left(\sum_{k\in \mathcal{K}} 4 h_k^2 b_k^2 + \left(\sum_{k\in \mathcal{K}} h_k b_k \right)^2 + n \sigma^2 \right) \label{e:ineq_smooth_strong_fourth} \\
\leq & \left( 1- \frac{2 M \eta^{(t)} a \sum_{k\in \mathcal{K}} h_k b_k \cos (\theta_{\text{th}}) }{\lVert \nabla F(\bm{w}^{(t)}) \rVert} \right) \mathbb{E} \left \{\lVert \bm{w}^{(t)} - \bm{w}^* \rVert^2 \right \}
+ \left(\eta^{(t)}\right)^2 a^2 \left(\sum_{k\in \mathcal{K}} 4 h_k^2 b_k^2 + \left(\sum_{k\in \mathcal{K}} h_k b_k \right)^2 + n \sigma^2 \right)  \label{e:ineq_smooth_strong_fifth}
\end{align}
\end{subequations}
where
\begin{itemize}
  \item the inequality in (\ref{e:ineq_smooth_strong_third}) holds because $\nabla F(\bm{w}^*) = 0$;
  \item the inequality  in (\ref{e:ineq_smooth_strong_fourth}) is established since there is
  $\left(\bm{w}^{(t)} - \bm{w}^{*} \right)^T \left(\nabla F(\bm{w}^{(t)}) - \nabla F(\bm{w}^*) \right) \geq M \lVert \bm{w}^{(t)} - \bm{w}^{*}\rVert^2$
   when $F(\bm{w})$ satisfies Assumption \ref{a:convexity} according to \cite{Nesterov};
  \item the inequality in (\ref{e:ineq_smooth_strong_fifth}) comes from the fact $\cos(\theta_k) \geq \cos(\theta_{\text{th}})$ for $k\in \mathcal{K}$ according to Assumption \ref{a:bias}.
\end{itemize}

Define
\begin{equation}
q^{(t)} \triangleq \left( 1- \frac{2 M \cos (\theta_{\text{th}})\eta^{(t)} a \sum_{k\in \mathcal{K}} h_k b_k}{ \lVert \nabla F(\bm{w}^{(t)}) \rVert } \right), \forall t = 1, 2, ...
\end{equation}
and set $\eta^{(t)}$ as $\eta$.
By selecting the $\eta$, $a$, $\{b_k| k\in \mathcal{K}\}$ such that $q^{(t)}<1$, which requires
\begin{equation} \label{e:a_requirement}
 0< 2 M \cos (\theta_{\text{th}}) \eta a \sum_{k\in \mathcal{K}} h_k b_k .
\end{equation}
The inequality in (\ref{e:a_requirement}) can be easily fulfilled by simply setting $\eta>0$, $a>0$, $\{b_k| k\in \mathcal{K}\}$ such that $\sum_{k\in \mathcal{K}} h_k b_k > 0$.
On the other hand, since $\lVert \nabla F(\bm{w}^{(t)})\rVert \leq G$ for any $\bm{w}^{(t)}$ according to Assumption \ref{a:gradient_bound}, there is
\begin{small}
\begin{equation}
q^{(t)} \leq  \left(1 - \frac{2M \cos (\theta_{\text{th}}) \eta a \sum_{k\in \mathcal{K}} h_k b_k}{G}\right)<1, \forall t=1, 2, ....
\end{equation}
\end{small}
With
\begin{equation*}
q^{\max} = \max \left(\left(1 - \frac{2M \cos (\theta_{\text{th}}) \eta a \sum_{k\in \mathcal{K}} h_k b_k}{G}\right), 0\right),
\end{equation*}
 which satisfies $q^{(t)} \leq q^{\max} <1$ for $t \geq 1$. Recalling the expression given in (\ref{e:ineq_smooth_strong}), there is
\begin{subequations} \label{e:ineq_smooth_strong_T_1}
\begin{align}
& \mathbb{E} \left\{ \lVert \bm{w}^{(T)} - \bm{w}^* \rVert^2 \right\} \notag \\
& \leq \prod_{t=1}^{T-1} q^{(t)} \mathbb{E} \left\{ \lVert \bm{w}^1 - \bm{w}^*\rVert^2 \right\} +
\left( \sum_{t=1}^{T-1} \left(\eta^{(T-t)}\right)^2 \prod_{i=1}^{t-1} q^{(T-i)} \right)
\times a^2 \left(\sum_{k\in \mathcal{K}} 4 h_k^2 b_k^2 + \left(\sum_{k\in \mathcal{K}} h_k b_k \right)^2 +  n \sigma^2 \right) \label{e:ineq_smooth_strong_T_1_a} \\
& \leq  \left(q^{\max}\right)^{T-1}  \mathbb{E} \left \{\lVert \bm{w}^1 - \bm{w}^*\rVert^2 \right\} +  \left(\sum_{t=1}^{T-1} \eta \left(q^{\max}\right)^{t-1}\right)
\times a^2 \left(\sum_{k\in \mathcal{K}} 4 h_k^2 b_k^2 + \left(\sum_{k\in \mathcal{K}} h_k b_k \right)^2 + n \sigma^2 \right) \label{e:ineq_smooth_strong_T_1_b} \\
& =  \left(q^{\max}\right)^{T-1}  \mathbb{E} \left\{ \lVert \bm{w}^1 - \bm{w}^*\rVert^2 \right\} +  \eta^2 \left(\frac{1 - \left(q^{\max}\right)^{T-1}}{1 - q^{\max}}\right)
\times a^2 \left(\sum_{k\in \mathcal{K}} 4 h_k^2 b_k^2 + \left(\sum_{k\in \mathcal{K}} h_k b_k \right)^2 + n \sigma^2 \right) \label{e:ineq_smooth_strong_T_1_c} \\
& \leq  \left(q^{\max}\right)^{T-1}  \mathbb{E} \left \{ \lVert \bm{w}^1 - \bm{w}^*\rVert^2 \right\}
+ \max\left(\frac{a \eta G}{2M \cos (\theta_{\text{th}}) \sum_{k\in \mathcal{K}} h_k b_k} ,a^2 \eta^2 \right)
\times \left(\sum_{k\in \mathcal{K}} 4 h_k^2 b_k^2 + \left(\sum_{k\in \mathcal{K}} h_k b_k \right)^2 + n \sigma^2 \right).\label{e:ineq_smooth_strong_T_1_d}
\end{align}
\end{subequations}

For the inequality in (\ref{e:ineq_smooth_strong_T_1}), it is worthy to mention that $q^{(t)}$ or even $q^{\max}$ may be no larger than zero for some $t \geq1$, say $t'$, which happens when
\begin{equation}
2 M \cos (\theta_{\text{th}}) \eta a \sum_{k\in \mathcal{K}} h_k b_k \geq \lVert \nabla F(\bm{w}^{(t')})\rVert.
\end{equation}
In this case, by setting $q^{(t')}$ to be zero, the inequality in (\ref{e:ineq_smooth_strong_T_1_b}) still holds, no matter $q^{\max}$ is equal to or larger than 0.

This completes the proof.

\section{Proof of Lemma \ref{lem:CaseI_convex_set}} \label{app:lemma3}
To prove this lemma, we only need to prove the left-hand side function of (\ref{e:CaseI_lower_convex_simple_a}), denoted as
\begin{equation*}
h(\{b_k | k\in \mathcal{K}\}) \triangleq \sqrt{ \left(\sum_{k\in \mathcal{K}} 4 h_k^2 b_k^2  + n \sigma^2 \right)},
\end{equation*}
which is convex with respect to the vector of $\{b_k | k \in \mathcal{K}\}$.
Define a $(K+1)$ dimensional vector $\bm{x} = \left(x_1, ..., x_K, x_{K+1}\right)^T$. Then a function with respect to $\bm{x}$,
\begin{equation}
f(\bm{x}) \triangleq  \lVert \bm{x} \rVert = \sqrt{\sum_{k=1}^{K+1} \left(x_k\right)^2 }
\end{equation}
can be found to be convex according to \cite{Boyd}, and the function
\begin{equation}
g(\bm{x}) \triangleq \sqrt{\sum_{k\in \mathcal{K}} 4 h_k^2 x_k^2  +  x_{K+1}^2}
\end{equation}
 is also a convex function with $\bm{x}$ since it can be taken as the composition of the convex function $f(\bm{x})$ with an affine mapping \cite{Boyd}. With the convexity of $g(\bm{x})$, for two vectors $\bm{x}^{\dag} =\left(b_1^{\dag}, b_2^{\dag}, ...., b_K^{\dag}, \sigma \right)^T$ and $\bm{x}^{\ddag} = \left(b_1^{\ddag}, b_2^{\ddag}, ...., b_K^{\ddag}, \sigma \right)^T$ and any $\theta \in [0,1]$, there is
\begin{equation} \label{e:g_convex}
g(\theta \bm{x}^{\dag} + (1-\theta) \bm{x}^{\ddag}) \leq \theta g(\bm{x}^{\dag}) + (1-\theta) g(\bm{x}^{\ddag}),
\end{equation}
then there is
\begin{equation} \label{e:convexity_ineq_proof}
\begin{array}{ll}
& \sqrt{\sum_{k\in \mathcal{K}} 4 h_k^2 \left(\theta b_k^{\dag} + (1-\theta) b_k^{\ddag}\right)^2  + n \sigma^2 } \\
 = &  \sqrt{\sum_{k\in \mathcal{K}} 4 h_k^2 \left(\theta b_k^{\dag} + (1-\theta) b_k^{\ddag}\right)^2  + \left(\theta \sigma + (1-\theta) \sigma \right)^2 } \\
\leq &  \theta \sqrt{ \left(\sum_{k\in \mathcal{K}} 4 h_k^2 \left(b_k^{\dag}\right)^2  + n \sigma^2 \right)} +(1-\theta)  \sqrt{ \left(\sum_{k\in \mathcal{K}} 4 h_k^2 \left(b_k^{\ddag}\right)^2  + n \sigma^2 \right)}
\end{array}
\end{equation}
where the inequality in (\ref{e:convexity_ineq_proof}) comes from the convexity of $g(\bm{x})$ shown in (\ref{e:g_convex}).

To this end, it can be seen that the inequality in (\ref{e:convexity_ineq_proof}) exactly shows that the function $h(\{b_k | k\in \mathcal{K}\})$ is convex with the vector of $\{b_k | k\in \mathcal{K}\}$.

This completes the proof.
\end{appendices}

\ifCLASSOPTIONcaptionsoff
  \newpage
\fi



\end{document}